\documentclass[10pt,twocolumn,letterpaper]{article}

\usepackage[accsupp]{axessibility}
\usepackage{wacv}
\usepackage{times}
\usepackage{epsfig}
\usepackage{graphicx}
\usepackage{amsmath}
\usepackage{amssymb}
\usepackage{graphicx}
\usepackage{caption}
\usepackage{subfigure}
\usepackage{color}
\usepackage{comment}
\usepackage{float}

\usepackage{balance}

\def\wacvPaperID{29} 

\wacvfinalcopy 

\ifwacvfinal
\fi

\ifwacvfinal
\usepackage[breaklinks=true,bookmarks=false]{hyperref}
\else
\usepackage[pagebackref=true,breaklinks=true,colorlinks,bookmarks=false]{hyperref}
\fi

\ifwacvfinal
\else
\fi

\begin{document}


\title{SAPNet: Segmentation-Aware Progressive Network for Perceptual Contrastive Deraining }

\author{Shen Zheng, Changjie Lu, Yuxiong Wu and Gaurav Gupta\\
Wenzhou-Kean University\\
Wenzhou, China\\
{\tt\small  {zhengsh, lucha, yuxiongw, ggupta}@kean.edu}



}


\maketitle 
\thispagestyle{empty}
\begin{figure*} 
\centering
\subfigure{
    \includegraphics[width=3.4cm]{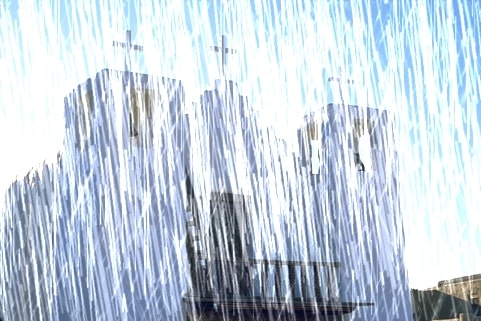}}\hspace*{-0.9em} 
\subfigure{
    \includegraphics[width=3.4cm]{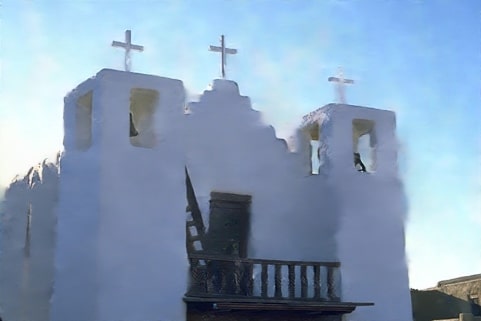}}\hspace*{-0.9em} 
\subfigure{    
    \includegraphics[width=3.4cm]{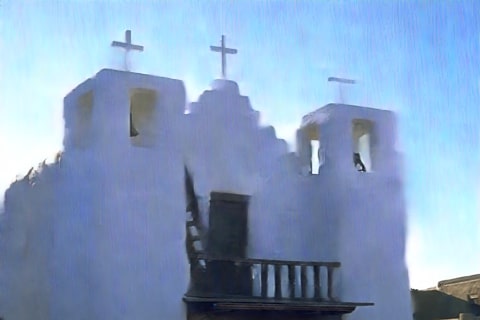}}\hspace*{-0.9em} 
\subfigure{
    \includegraphics[width=3.4cm]{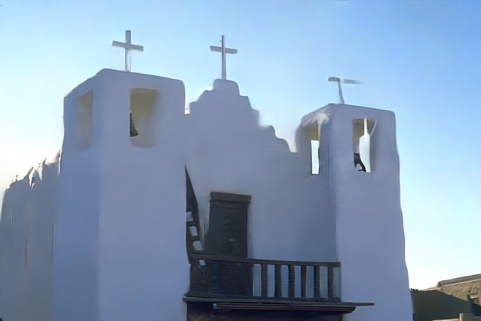}}\hspace*{-0.9em} 
\subfigure{
    \includegraphics[width=3.4cm]{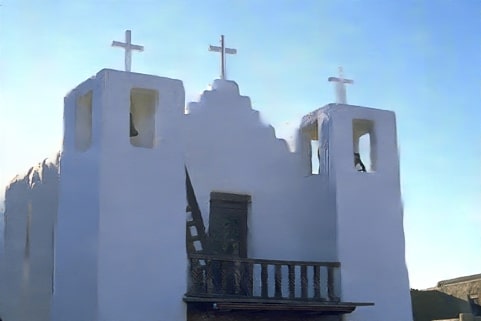}}\hspace*{-0.9em} 
    \\
    \vspace*{-0.9em}
\subfigure{
    \includegraphics[width=3.40cm]{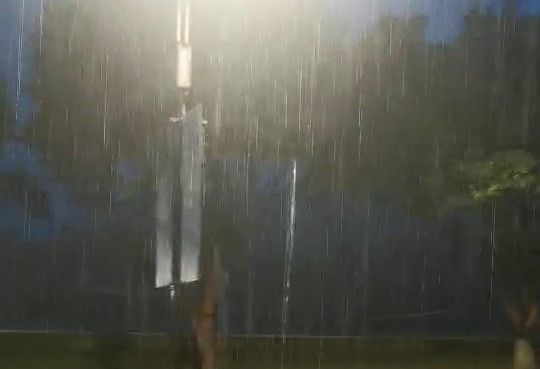}}\hspace*{-0.9em} 
\subfigure{
    \includegraphics[width=3.4cm]{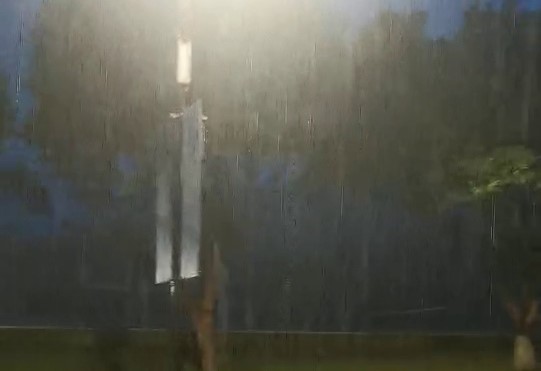}}\hspace*{-0.9em} 
\subfigure{
    \includegraphics[width=3.42cm]{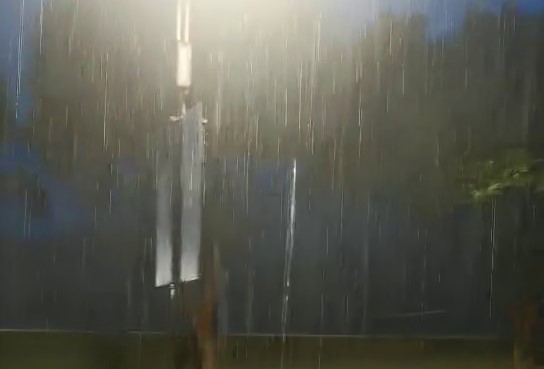}}\hspace*{-0.9em} 
\subfigure{
    \includegraphics[width=3.38cm]{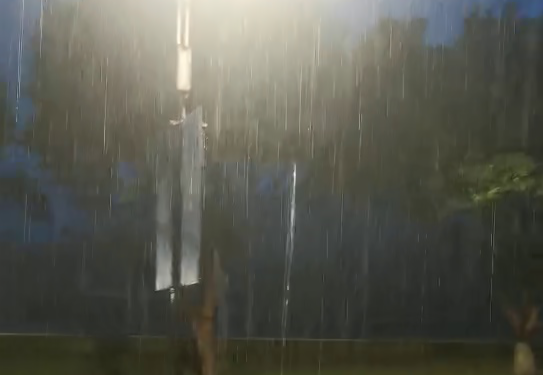}}\hspace*{-0.9em} 
    \subfigure{
    \includegraphics[width=3.4cm]{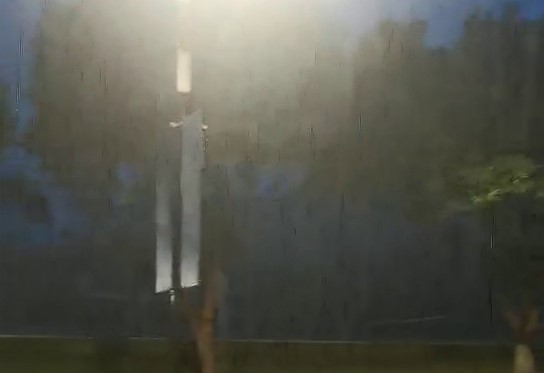}}\hspace*{-0.9em} 
    \caption{Deraining comparison at a synthetic rainy image from Rain100H (top) and a real rainy image (bottom). From left to right: Rainy, PreNet (CVPR 2019), MSPFN (CVPR 2020), MPRNet (CVPR 2021), SAPNet (ours). Compared with previous state-of-the-arts, the proposed model is superior at rain removal, edge preservation, blur suppression and color balance. Our advantage is more evident when it comes to real rainy images under complex illumination conditions.}
    \label{Rain100H(1)}
\end{figure*}




\begin{abstract}

Deep learning algorithms have recently achieved promising deraining performances on both the natural and synthetic rainy datasets. As an essential low-level pre-processing stage, a deraining network should clear the rain streaks and preserve the fine semantic details. However, most existing methods only consider low-level image restoration. That limits their performances at high-level tasks requiring precise semantic information. To address this issue, in this paper, we present a segmentation-aware progressive network (SAPNet) based upon contrastive learning for single image deraining. We start our method with a lightweight derain network formed with progressive dilated units (PDU). The PDU can significantly expand the receptive field and characterize multi-scale rain streaks without the heavy computation on multi-scale images. A fundamental aspect of this work is an unsupervised background segmentation (UBS) network initialized with ImageNet and Gaussian weights. The UBS can faithfully preserve an image's semantic information and improve the generalization ability to unseen photos. Furthermore, we introduce a perceptual contrastive loss (PCL) and a learned perceptual image similarity loss (LPISL) to regulate model learning. By exploiting the rainy image and groundtruth as the negative and the positive sample in the VGG-16 latent space, we bridge the fine semantic details between the derained image and the groundtruth in a fully constrained manner. Comprehensive experiments on synthetic and real-world rainy images show our model surpasses top-performing methods and aids object detection and semantic segmentation with considerable efficacy. A Pytorch Implementation is available at https://github.com/ShenZheng2000/SAPNet-for-image-deraining.
\vspace{2.9em}
\end{abstract}

\begin{figure}[t]
\centering
\includegraphics[width=8.5cm]{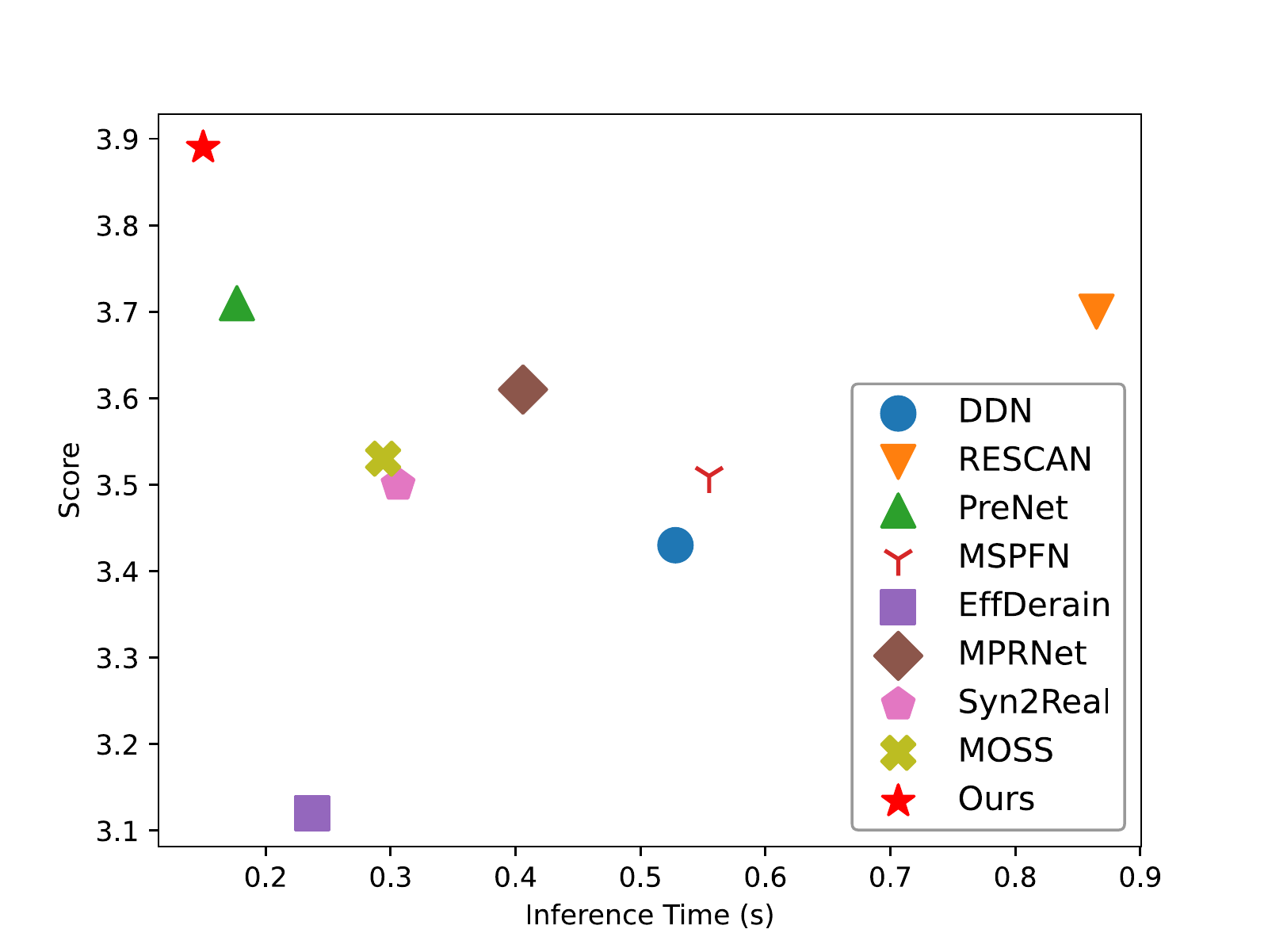}
\caption{User study score$\uparrow$ and inference time$\downarrow$ comparison. The user study score (1-5) is averaged from real-rainy datasets including Rain800, SIRR and MOSS. The average run time (inference time) is calculated on images with size 512 $\times$ 512 with a single NVIDIA RTX 2080 Ti GPU.}
\label{TimeEfficiency}
\end{figure}

\section{Introduction}
\footnote{This work is supported by the research funding from
Wenzhou-Kean University with grant SpF2021011.} Rain is typical weather that degrades the visibility of images and videos. Especially in heavy rain, the combination of rain streaks and accumulation has a severe adverse impact on computer vision tasks, such as image classification, object detection, and semantic segmentation \cite{li2019single}. Therefore, it is crucial to remove rains and to recover the rainy images. Since 2017, deep learning deraining methods, based upon CNN \cite{fu2017removing, yang2017deep, li2018recurrent, ren2019progressive, jiang2020multi, qian2018attentive}, or GAN \cite{goodfellow2014generative, qian2018attentive, pan2020physics}, have attracted significant attention due to their outstanding accuracy, capacity, and flexibility. 

Despite their progress in benchmark datasets, both directions focus on image quality scores like MSE/MAE and fail to consider whether their rain removal benefits high-level vision tasks such as detection and segmentation. Indeed, it has been shown by \cite{haris2018task, pei2018does,li2019single} that only considering image quality metrics does not guarantee better performance at advanced tasks. Motivated by that observation, recent models \cite{liu2017image, haris2018task, zheng2021deblur} have explored joint training to bridge the gap between low-level and high-level tasks. However, those approaches require heavy amounts of annotated images. The acquisition of those data requires tedious manual labelling, which is expensive and time-consuming. The synthetically generated labelled image also easily overfit a model, therefore compromising the generalization to real-world images.

One common way to improve the generalization ability is to transfer the knowledge from the synthetic rain domain to the real rain domain, using methods like Gaussian mixture model \cite{wei2019semi},  Gaussian process \cite{yasarla2020syn2real}, and self-supervised memory block \cite{huang2021memory}. Although these strategies improve real-world images that have consistent rain patterns, those deraining methods face significant performance degradation with heavy/dense rain streaks due to their failure to characterize the information from different scales and magnitudes. On the other hand, multi-scale deraining methods \cite{zhang2019image, jiang2020multi, guo2020efficientderain} require accumulating model parameters to address images resized to different scales. Consequently, the long inference time (Fig. \ref{TimeEfficiency}) and the growing model size restrict their deployment on mobile devices or real-time deraining applications like autonomous driving and surveillance. 

\begin{figure*}
\centering
\includegraphics[width=17cm]{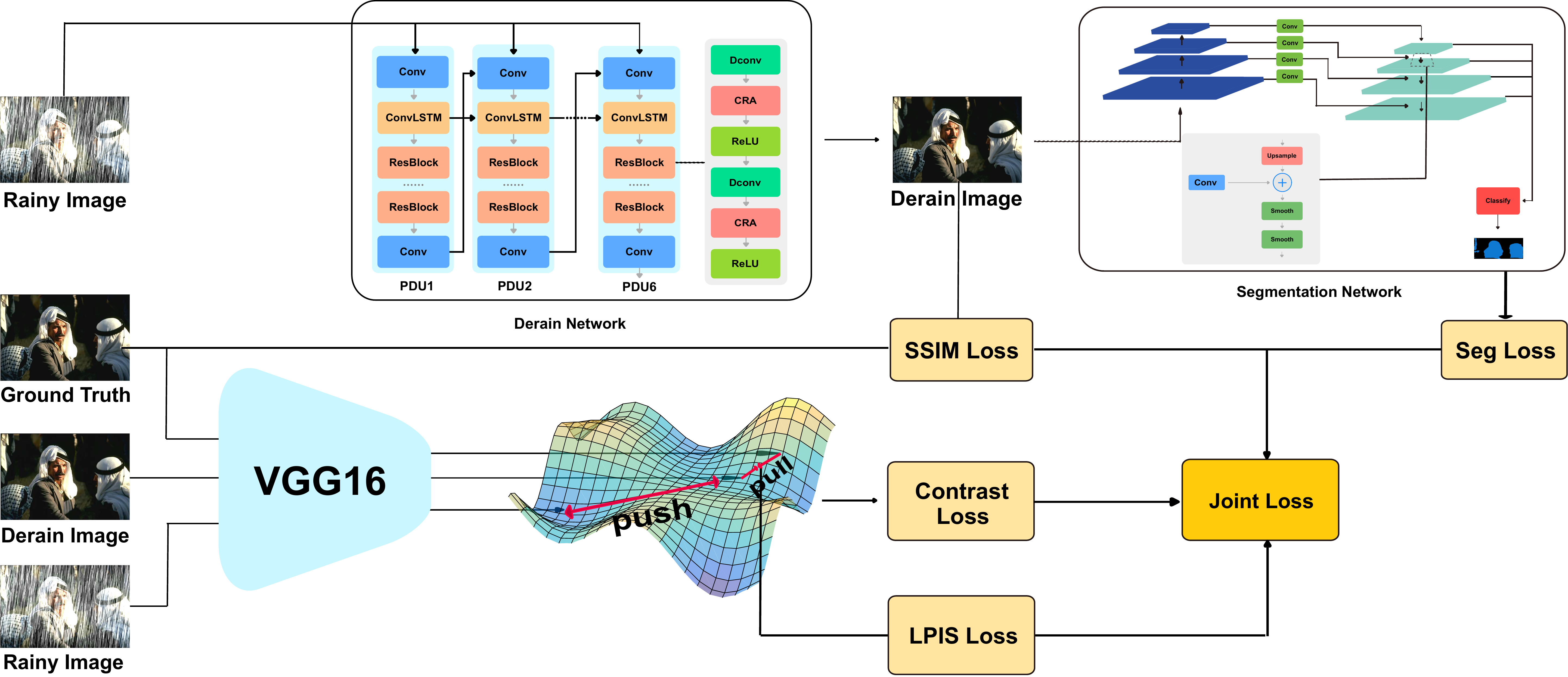}
\caption{Model architecture for SAPNet. SAPNet joins a derain network for supervised rain removal, a segmentation network for unsupervised background segmentation, and a VGG-16 network for perceptual contrast. The rainy image first enters the derain network for rain removal, using the groundtruth as the reference to obtain the negative ssim loss. Once the deraining process is finished, the segmentation network will consume the derained image to calculate the segmentation loss. Meanwhile, the perceptual contrastive loss and the learned perceptual image similarity loss will be computed on the VGG16 latent space using the rainy image, the derained image, and the groundtruth. Finally, the jointed loss will update the derain network during training.  Here we use `DConv' for dilated convolution and `CRA' for the channel residual attention block in Fig. \ref{ModelBlocks}}
\label{ModelArchs}
\end{figure*}

To address the limitations of previous researches, we propose SAPNet, a \textbf{s}egmentation-\textbf{a}ware \textbf{p}rogressive \textbf{net}work for single image deraining (Fig. \ref{ModelArchs}). Due to the importance of multi-scale contextualized rain streaks information in removing heavy/dense rains, we first introduce a progressive dilated network to expand the receptive field significantly and reuse the previous recurrent stage's knowledge without additional parameters. As the semantic information is essential for task-driven deraining, we then present an unsupervised background segmentation network to preserve the semantic details during rain removal without segmentation label. Inspired by the success of contrastive learning and perceptual similarity in low-level vision tasks, we also exploit the rainy images as the negative samples to guide rain removal. 

The contribution of this paper can be highlighted as four folds:
\begin{itemize}
    \item We propose a segmentation-aware progressive network for single image deraining. To the best of our knowledge, we are first to utilize unsupervised background segmentation to aid rain removal.
    \item We present a novel progressive network formed with progressive dilated units (PDU). This design allows an efficient usage of multi-scale rain streak information.
    \item We design a new perceptual contrastive loss (PCL) and a learned perceptual image similarity loss (LPISL). With the advantage of contrastive learning and perceptual similarity, the derained image is close to the groundtruth in terms of pixel-wise difference and fine details.
    \item Comprehensive experiments demonstrate that our model surpasses previous state-of-the-arts qualitatively and quantitatively. 
\end{itemize}

\section{Related Work}
In this section, we display a brief review on deep learning-based image deraining methods and contrastive learning-based image restoration approaches.

\noindent
\subsection{Deep Learning for Single Image Deraining}
Deep Learning methods have demonstrated excellent performance in rain removal. For instance, DetailNet \cite{fu2017removing} uses a prior-based deep detail network to estimate rain streaks with negative residual information. Jorder \cite{yang2017deep} utilizes a multi-task architecture to learn binary rain streak maps, heavy rain streaks appearance, and the clean background in one go. 

Recurrent networks have been employed to construct more efficient image deraining models. For example, RESCAN \cite{li2018recurrent} leverages a recurrent neural network with squeeze-and-excitation blocks for rain removal. PreNet \cite{ren2019progressive} recursively unfolds a shallow residual network to process the input and intermediate layers progressively. ID-CGAN \cite{qian2018attentive} proposed an attention-based GAN \cite{goodfellow2014generative} for attending raindrops and its neighbor backgrounds. 

Recent deraining methods \cite{yasarla2019uncertainty, jiang2020multi, zamir2021multi} have begun to corporate multi-scale learning to exploit rain streaks of different sizes and directions. For instance, MSPFN \cite{jiang2020multi} utilizes a multi-scale pyramid architecture to supervise the fine fusion of rain streaks information. 

Different from previous approaches, we utilize dilated convolution \cite{yu2015multi} to expand the receptive field. In this way, we obtain rain streaks of different scales within one recurrent unit without compromising the computational efficiency. We also exploit unsupervised semantic segmentation to restore the background semantic details during intensive rain removal.

\noindent
\subsection{Contrastive Learning for Image Restoration}
Contrastive Learning have made notable progress in self-supervised representation learning \cite{chen2020simple, he2020momentum, henaff2020data}. The goal of contrastive learning is to pull an anchored sample near to the positive sample and, meanwhile, push that anchored sample away from the negative sample in the given latent space. Previous contrastive learning often targets high-level vision tasks like image classification and object detection. 

Recently, contrastive learning has been utilized in low-level vision tasks. For example, \cite{park2020contrastive} has shown that contrastive learning can boost the performance of unpaired image-to-image translation. Contrastive Learning is also applied in image dehazing \cite{wu2021contrastive} with a pixel-wise L1 loss and in image super-resolution \cite{wang2021towards} with self-supervised knowledge distillation.

Unlike previous contrastive learning methods, this work applies contrastive learning to single image deraining for the first time. To better reserve the fine details in a photo during rain removal, we take perceptual similarity into account and present a new perceptual contrastive loss.

\begin{figure}[t]
\centering
\includegraphics[width=8cm]{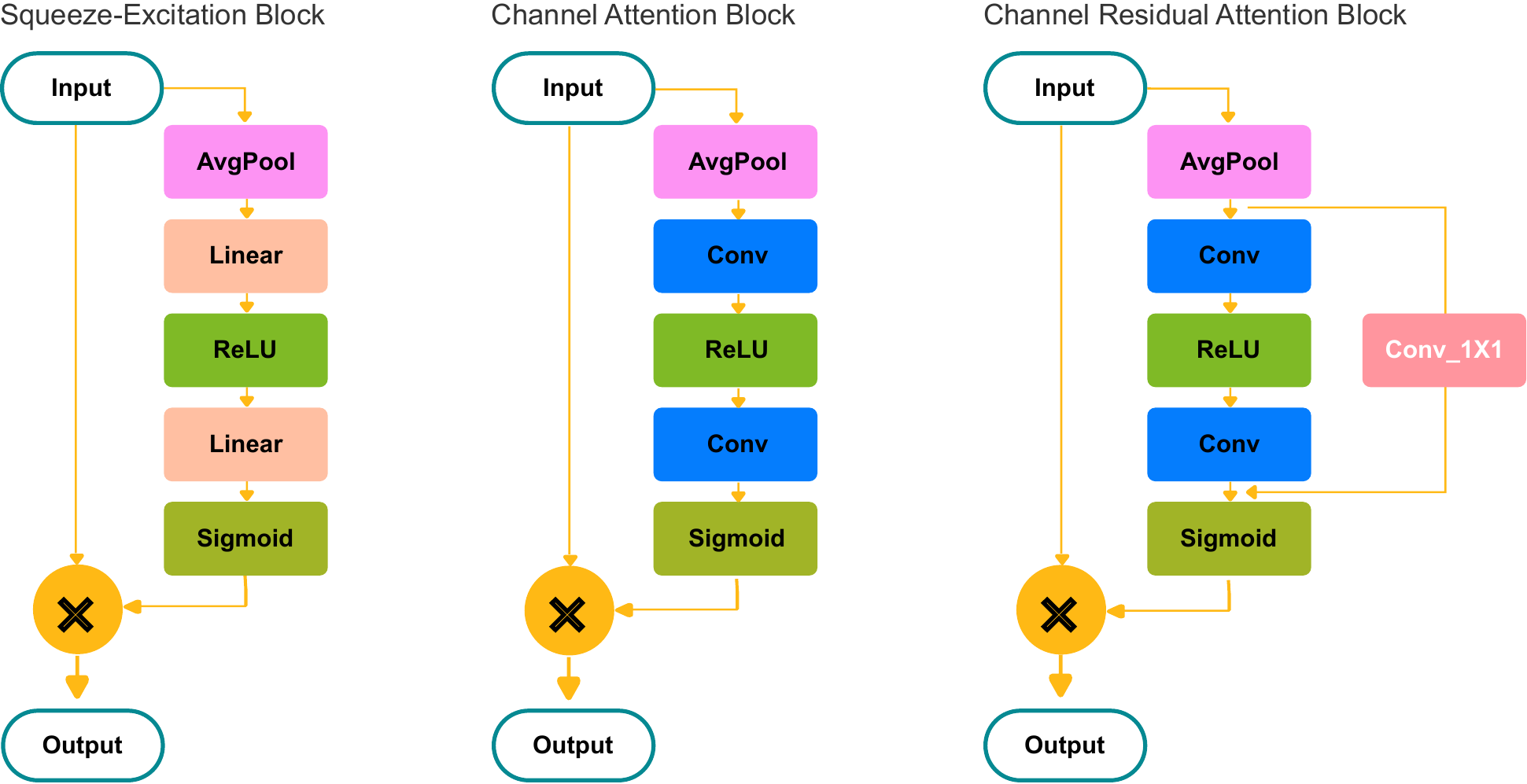}
\caption{Different attention blocks for image deraining. From left to right: squeeze-excitation (SE) block, channel attention (CA) block, and the proposed channel residual attention (CRA) block. }
\label{ModelBlocks}
\end{figure}

\section{Methodology}
In this section, we present the proposed SAPNet by analyzing the building components, network architecture, and loss functions. 

\subsection{Channel Residual Attention Block}
Building blocks are essential for rain removal because they determine a model’s ability to characterize the rain streak patterns. Recently, state-of-the-arts deraining methods \cite{li2018recurrent, qian2018attentive, jiang2020multi, zamir2021multi} have begun to incorporate the attention mechanism to boost deraining performances. In Fig. \ref{ModelBlocks}, we display three effective blocks for image deraining, including squeeze-excitation (SE) block \cite{hu2018squeeze}, channel attention (CA) block \cite{woo2018cbam} and our proposed channel residual attention (CRA) block. Compared with SE and CA, our skip connection from the pooling layer to the sigmoid activation function allows a more efficient feature fusion and gradient flow during the model training. The ablation study will analyze the superiority of CRA qualitatively and quantitatively.


\begin{figure}[t]
\centering
\includegraphics[width=8cm]{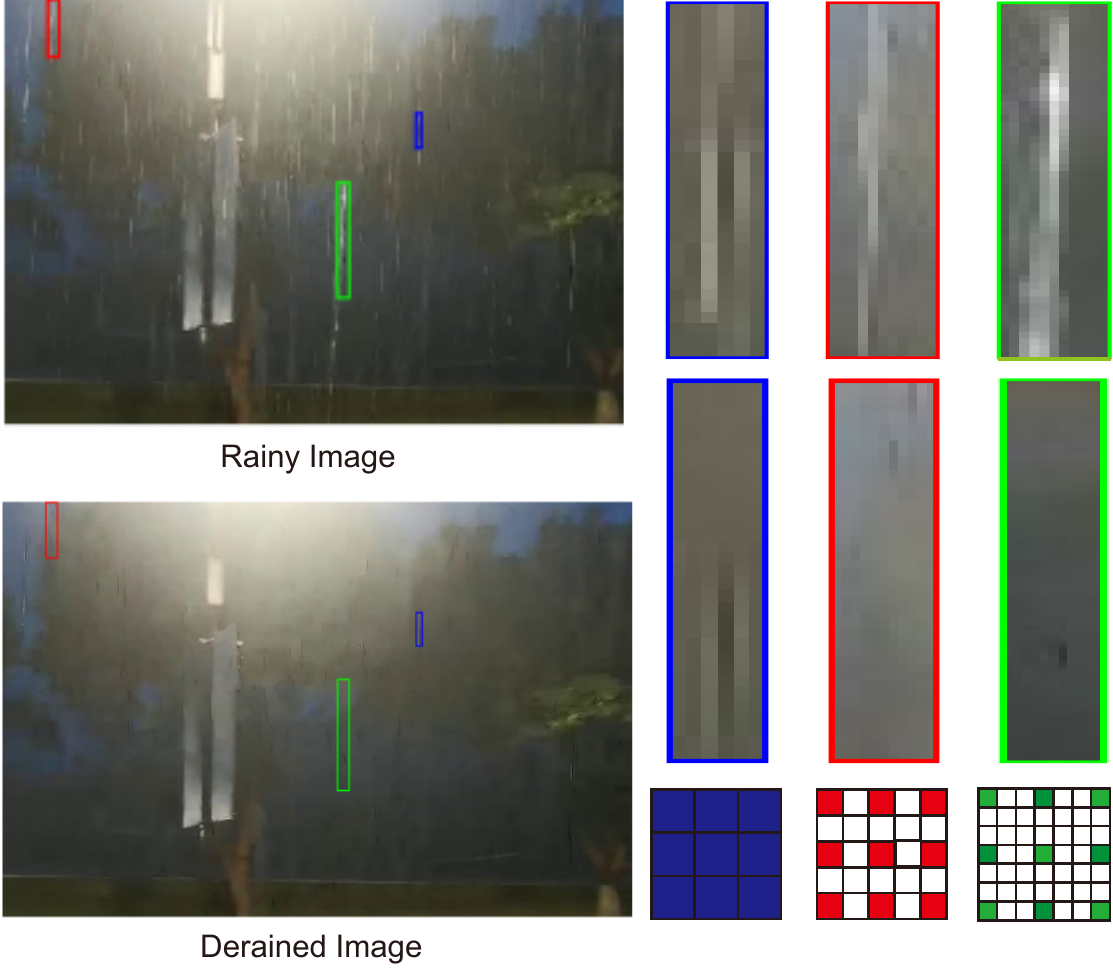}
\hspace*{-0.9em}
\caption{Visual Illustration of Progressive Dilation. We use blue, red, green bounding boxes to highlight rain streaks of diverse shapes and thickness. In the proposed network, each progressive dilated unit utilize convolutions with different dilation rate to capture and clear multi-scale contextualized rain streaks. }
\label{PiexedDilation}
\end{figure}

\subsection{Progressive Dilated Unit}
A basic neural network like \cite{fu2017removing} cannot characterize heavy/dense rain streaks, as shown by \cite{ren2019progressive, zamir2021multi}.



Inspired by the success of progressive networks \cite{li2018recurrent, ren2019progressive} for image deraining and the efficiency of dilated convolution for multi-scale information \cite{chen2017deeplab, chen2017rethinking} , we design a progressive dilated unit (PDU), which uses dilated convolution to exploit the multi-scale contextualized rain streaks information (See Fig. \ref{PiexedDilation}). Our PDU contains four parts, a leading convolutional block for consuming the input rainy image, five proposed residual blocks for feature extraction and an ending convolution block for yielding the derained image. The dilation rate of the residual blocks are 1, 2, 4, 8, 16, respectively. 

Each proposed residual block includes, in a single repetition, a convolution layer, a channel attention block, and a ReLU \cite{nair2010rectified} activation layer. The channel number of the convolution layers is 32, and the kernel size is 3. Besides, the reduction factor of the CRA is 16. Finally, we have an output convolution layer that reduces the channel number from 32 to 3.

\subsection{Derain Network}
Our derain network consists of 6 recurrent stages. Each stage corresponds to a progressive dilated unit (PDU) which has shared parameters with others. The inference of our network is:

\begin{equation}
\begin{array}{l}
\mathbf{s}^{t}=f_{\text {rec }}\left(\mathbf{s}^{t-1}, f_{\text {in }}\left(\mathbf{x}^{t-1}, \mathbf{y}\right)\right) \\
\mathbf{x}^{t}=f_{\text {out }}\left(f_{\text {res }}\left(\mathbf{s}^{t}\right)\right)
\end{array}
\end{equation}
where $\mathbf{y}$ is the rainy image. where $f_{\text {in}}$ and $f_{\text {out}}$ is the convolutional block for receiving the input and outputting the results, respectively. $f_{\text {rec}}$ is the recurrent operations. $f_{\text {res}}$ is the residual blocks. $\mathbf{s}^{t}$ is the recurrent state at stage $t$. $\mathbf{x}^{t}$ is the derained image at stage $t$. Note that we combine the rainy image and the derained image from previous recurrent unit as the input for the next recurrent unit. That strategy is shown by \cite{qian2018attentive, ren2019progressive} to boost deraining performance.

For the recurrent calculations, we leverage convolutional LSTM \cite{xingjian2015convolutional} for a more consistent cross-stage interaction. It can be formulated as:

\begin{equation}
\begin{array}{l}
i_{t}=\sigma\left(W_{x i} * x_{t}+W_{h i} * h_{t-1}+W_{c i} \circ c_{t-1}+b_{i}\right) \\
f_{t}=\sigma\left(W_{x f} * x_{t}+W_{h f} * h_{t-1}+W_{c f} \circ c_{t-1}+b_{f}\right) \\
c_{t}=f_{t} \circ c_{t-1}+i_{t} \circ \tanh \left(W_{x c} * x_{t}+W_{h c} * h_{t-1}+b_{c}\right) \\
o_{t}=\sigma\left(W_{x o} * x_{t}+W_{h o} * h_{t-1}+W_{c o} \circ c_{t}+b_{o}\right) \\
h_{t}=o_{t} \circ \tanh \left(c_{t}\right)
\end{array}
\end{equation}
where $i_{t}$ is the input gate, $f_{t}$ is the forget gate, $o_{t}$ is the output gate $c_{t}$ and is the cell state, $\circ$ denotes the element-wise product, and $*$ denotes the convolution operation.

\subsection{Segmentation Network}
Semantic segmentation has been useful for low-level vision tasks such as image denoising \cite{liu2017image, liu2020connecting}, image deblurring \cite{ren2017video} and image deraining \cite{zhang2020beyond}. Inspired by this, we design an unsupervised background segmentation network that performs semantic segmentation on the derained image. Similar to \cite{liu2017image, wang2019segmentation}, we freeze the parameters of the entire segmentation network during training. 

Motivated by the success of the feature pyramid network \cite{lin2017focal} in utilizing multi-scale contextual information, which is essential for image deraining, our segmentation network uses an FPN backbone which consists of an encoder-decoder framework with lateral connections embedded with 1 $\times$ 1 convolution layer. Our encoder (bottom-up pathway) use ResNet-101 \cite{he2016deep} pretrained on ImageNet \cite{deng2009imagenet}, whereas our decoder (top-down pathway) is initialized with Gaussian weight with zero mean and a standard deviation of 0.05. Besides, both the encoder and the decoder have four convolution blocks. 

For each decoder stairs, the output image is bilinearly upsampled and concatenated with the lateral results. Two smooth layers of 3 $\times$ 3 convolution are designed for better perceptual quality after each concatenation. Finally, all stairs' image in the decoder is concatenated. The concatenated output's channel number is reduced from 512 to $n$, leading to a pixel-wise classification task with $n$-class. Empirically, we set $n$ to 21 because the benchmark image classification dataset PASCAL-VOC 2012 \cite{everingham2010pascal} have 20 significant object classes and another background class.




\subsection{Loss Function}

\begin{table}[t]
\centering
\begin{tabular}{l| l| l}
\hline
Name         & Category  & Test Samples \\ \hline
Rain12       & Synthetic & 12           \\ 
Rain100L    & Synthetic & 100          \\
Rain100H    & Synthetic & 100          \\
Rain800       & Real      & 50           \\ 
SIRR      & Real      & 147          \\
MOSS        & Real      & 48           \\ 
COCO150      & Synthetic & 150          \\
CityScape150 & Synthetic & 150          \\ \hline
\end{tabular}
 \vspace{0.4em}
\caption{Dataset Description}
\label{Dataset}
\end{table}

\noindent 
\textbf{Negative SSIM Loss}
Most single image deraining tasks use L2 loss for training. As shown by \cite{fu2019lightweight, ren2019progressive, wang2019spatial, jiang2020multi}, the L2 loss produces over-smoothed backgrounds and ghost artifacts, which is detrimental to the semantic information. As an alternative, we adopt negative SSIM loss to focus on luminance, contrast, and structure. The negative SSIM loss is:

\begin{equation}
\mathcal{L}_{\mathrm{ssim}}=-\operatorname{SSIM}\left(\mathbf{x}^{D}, \mathbf{x}^{G}\right)
\end{equation}
where $\mathbf{x}^{D}$, $\mathbf{x}^{R}$, and $\mathbf{x}^{G}$ represents the derained image, the rainy image, and the groundtruth, respectively.  is the rainy image.

\noindent 
\textbf{Segmentation Loss}
For unsupervised semantic segmentation of UBS, We utilize focal loss \cite{lin2017focal} to address the imbalance for rain streaks of different directions and magnitudes. Since the segmentation label is unavailable, we minimize the average of the cost function to make an overall more `confident' prediction. The segmentation loss is:

\begin{equation}
\mathcal{L}_{\mathrm{seg}}=\frac{1}{H W} \sum_{1 \leq i \leq H, 1 \leq j \leq W}-\alpha\left(1-p_{i, j}\right)^{\gamma} \log p_{i, j}
\end{equation}
where $H,W$ is the height and the width of the image. $p_{i, j}$ is the model's estimated probability for the class with a specific pixel-wise class probability in segmentation. Here we set $\alpha$ equals to 1 and $\gamma$ as 2.


\noindent 
\textbf{Perceptual Contrastive Loss}
A simple contrastive loss is usually based upon L1 loss \cite{park2020contrastive, wu2021contrastive, wang2021towards}. However, it is shown by \cite{johnson2016perceptual} that simple pixel-wise loss (L1/L2 loss) fails to reserve fine details and textures during image processing. Inspired by the success of perceptual loss \cite{johnson2016perceptual} in low-level vision tasks like image-to-image traslation \cite{isola2017image}, image super-resolution \cite{ledig2017photo}, and image deblurring \cite{kupyn2018deblurgan}, we inject perceptual loss into contrastive loss. The proposed perceptual contrastive loss is:

\begin{equation}
\mathcal{L}_{\mathrm{pcl}} = \sum_{i=1}^{n} \omega_{i} \cdot \frac{L1\left(V_{i}(\mathbf{x}^{D}), V_{i}(\mathbf{x}^{G})\right)}{L1\left(V_{i}(\mathbf{x}^{D}), V_{i}(\mathbf{x}^{R})\right)}
\end{equation}
where $V_{i}$ represents the $i^{th}$ extracted layer in from VGG-16. $\omega_{i}$ represents the weight coefficient to balance between shallow and deep layer features.

\noindent 
\textbf{Learned Perceptual Image Similarity Loss}
Learned Perceptual Image Patch Similarity (LPIPS) is first proposed in \cite{zhang2018unreasonable} to evaluate the perceptual similarity between the distorted image and the groundtruth. In this paper, we use the resized whole image rather than the cropped image patches proposed by the original paper. There are two reasons for this change. First, operating on the whole image helps restore the high-level semantic information crucial for detection and segmentation \cite{zheng2021semantic}. Second, perceptual similarity on the entire image explores non-local information \cite{wang2018non}, thereby complementing the convolution operations which can only process one local region at a time. We name our loss Learned Perceptual Image Similarity Loss (LPISL). The formulation is as below:

\begin{equation}
\mathcal{L}_{\text {lpisl}} =\sum_{i=1}^{n} \frac{1}{H_{i} W_{i}} \sum_{h, w}\left\|\theta_{i} \odot\left(V_{i}(\mathbf{x}^{D})-V_{i}(\mathbf{x}^{G}\right)\right\|_{2}^{2}
\end{equation}
where $\theta_{i}$ represents the cosine distance calculation.



\noindent 
\textbf{Total Loss}
Our total loss for SAPNet is:

\begin{equation}
\mathcal{L}=\lambda_{1} \times \mathcal{L}_{\mathrm{ssim}}+
\lambda_{2} \times \mathcal{L}_{\mathrm{seg}}+
\lambda_{3} \times \mathcal{L}_{\mathrm{pcl}}+
\lambda_{4} \times \mathcal{L}_{\mathrm{lpisl}}
\end{equation}
Here we set $\lambda_{1}$ to 1, $\lambda_{2}$, $\lambda_{3}$ and $\lambda_{4}$ to 0.1

\begin{figure}[t]
\centering
\includegraphics[width=2.0cm]{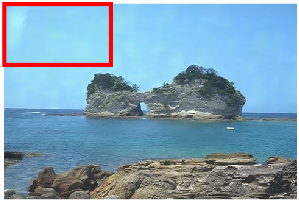}
\includegraphics[width=2.0cm]{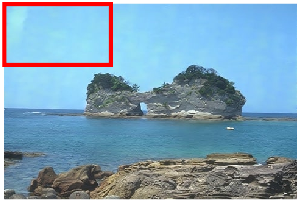}
\includegraphics[width=2.0cm]{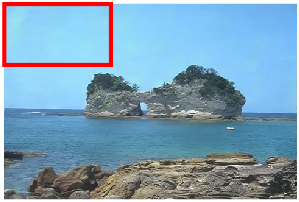}
\includegraphics[width=2.0cm]{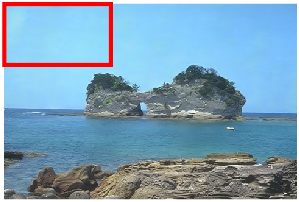} \\
\includegraphics[width=2.0cm]{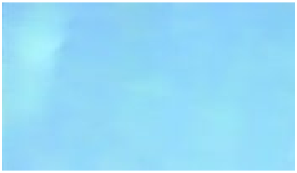}
\includegraphics[width=2.0cm]{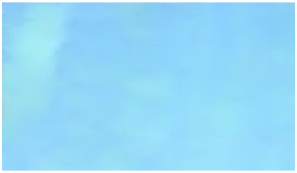}
\includegraphics[width=2.0cm]{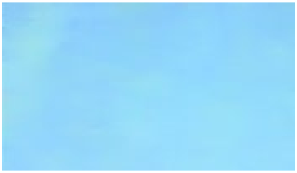}
\includegraphics[width=2.0cm]{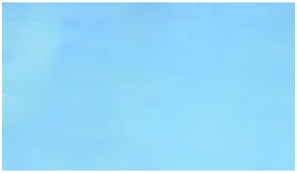}
\caption{Visual ablation of attention blocks. Top row: original image. Bottom row: cropped image. From left to right: Model-Conv, Model-SE, Model-CA, Model-CRA.}
\label{VisAttention}
\end{figure}

\begin{figure}[t]
\centering
\includegraphics[width=2.0cm]{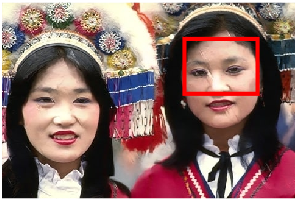}
\includegraphics[width=2.0cm]{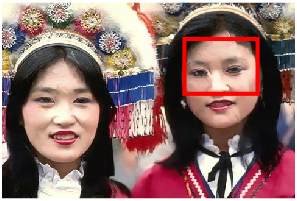}
\includegraphics[width=2.0cm]{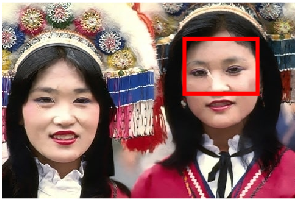}
\includegraphics[width=2.0cm]{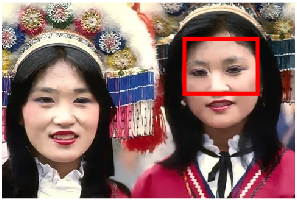} \\
\includegraphics[width=2.0cm]{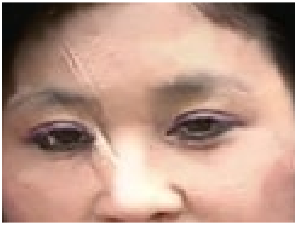}
\includegraphics[width=2.0cm]{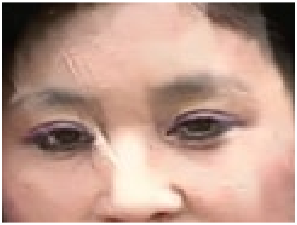}
\includegraphics[width=2.0cm]{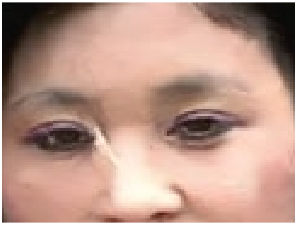}
\includegraphics[width=2.0cm]{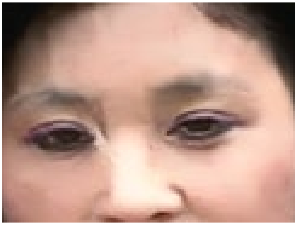} \\
\caption{Visual ablation of model components. Top row: original image from Model 1 to 4. Bottom row: cropped image from Model 1 to 4.}
\label{VisModel}
\end{figure}

\begin{figure}[t]
\centering
\includegraphics[width=2.0cm]{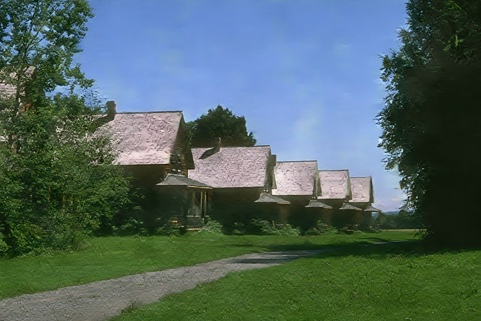}
\includegraphics[width=2.0cm]{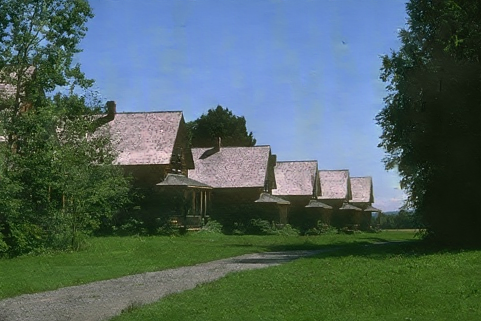}
\includegraphics[width=2.0cm]{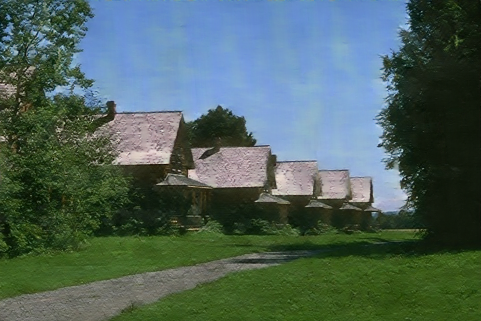}
\includegraphics[width=2.0cm]{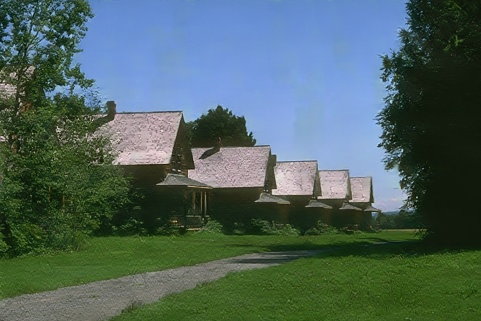}
\caption{Visual ablation of contrastive losses. From left to right, PreNet (No-CL), SAPNet (No-CL), SAPNet (L1-CL), SAPNet (PCL)}
\label{VisContrast}
\end{figure}

\section{Experiments}
\subsection{Implementation Details}
The proposed model is trained using Pytorch \cite{paszke2019pytorch} with one Tesla V100 GPU. The training dataset RainTrain100H \cite{yang2017deep} contains 1800 pairs of synthetic rainy images and the corresponding ground truth. The proposed model uses Adam \cite{kingma2014adam} optimizer for 100 epochs with an initial learning rate of 0.001 and a batch size of 12. The learning rate is reduced by 80 \% on epochs 30, 50, and 80, respectively. It takes around 20 hours for the model to converge.

\subsection{Datasets}
We utilize benchmark synthetic and real-world rainy datasets for comparisons. To investigate task-driven image deraining, we additionally chose 300 images in total from Microsoft COCO \cite{lin2014microsoft} and CityScape \cite{cordts2016cityscapes}. We synthesize rain for them and name the dataset COCO150 and CityScape150, respectively. Dataset details is at Table \ref{Dataset}.




\begin{table}[t]
\centering
\small
\begin{tabular}{l| l l l}
\hline
      & Model-SE & Model-CA & Model-CRA \\ \hline
PSNR & 28.74    & 28.89   & \textbf{29.46}           \\ 
SSIM  & 0.890 &  0.892  & \textbf{0.897}          \\ 
RT   & 0.166 & \textbf{0.149} & 0.150              \\\hline
\end{tabular}
\\  \vspace{0.4em}
\caption{Ablation on attention blocks in terms of PSNR$\uparrow$, SSIM$\uparrow$ and Run Time$\downarrow$}
\label{Aba_attention}
\end{table}

\begin{table}[t]
\small
\centering
\begin{tabular}{l|lll|lll}
\hline
       & \multicolumn{3}{c|}{Contrastive Loss} & \multicolumn{3}{l}{Prop. of Train Images} \\ \hline
Metric & No-CL       & L1-CL      & PCL        & 40 \%           & 60 \%          & 100  \%       \\ \hline
PSNR   & 28.96       & 26.84      & \textbf{29.46}      & 26.49        & 27.37        & \textbf{29.46}       \\ 
SSIM   & 0.888       & 0.853      & \textbf{0.897}      & 0.853        & 0.866        & \textbf{0.897}       \\ \hline
\end{tabular}
\caption{Ablation result for SAPNet with different contrastive lose and limited training images}
\label{CompareContrastLimit}
\end{table}

\subsection{Metrics}
Several benchmark metrics have been adopted for assessment. For the synthetic dataset, we use scikit-learn \cite{pedregosa2011scikit} as a unified library\footnote{Most papers use Matlab for computing their PSNR and SSIM. We find that, holding everything else fixed, the PSNR calculated from sklearn will be 1-2 db lower than the result from Matlab.} for PSNR and SSIM. For the real-world dataset, we use non-reference metrics, including UNIQUE \cite{zhang2021uncertainty} and BRISQUE \cite{mittal2012no}. For the task-driven parts, we use mean average precision (mAP) for object detection, mean pixel accuracy (mPA) and mean intersection over union (mIOU) for semantic segmentation. The method with the best and the second-best score is in \textbf{bold} and \underline{underline}, respectively.

\subsection{Baselines}
We compare the proposed method with recent state-of-the-arts. The superivsed method for comparison includes DDN \cite{fu2017removing}, RESCAN \cite{li2018recurrent}, PreNet \cite{ren2019progressive}, MSPFN \cite{jiang2020multi}, MPRNet \cite{zamir2021multi}, and EffDerain \cite{guo2020efficientderain}. The unsupervised method includes Syn2Real \cite{yasarla2020syn2real} and MOSS \cite{huang2021memory}. To ensure a fair comparison, all supervised methods for comparison is trained on RainTrain100H without data augmentation, using their publicly available codes.

\begin{table}[t]
\centering
\small
\begin{tabular}{l| l l l l l l}
\hline
         & M1 & M2 & M3 & M4 & M5 & Ours \\ \hline
CRA      &  \checkmark & \checkmark & \checkmark & \checkmark & \checkmark  & \checkmark  \\ 
UBS      &            & \checkmark  & \checkmark & \checkmark & \checkmark & \checkmark  \\
PCL      &            &            & \checkmark & \checkmark &  \checkmark  & \checkmark     \\
Dilation &            &            &            & \checkmark &  \checkmark  & \checkmark    \\
Decay    &            &            &            &            & \checkmark  & \checkmark  \\ 
LPISL    &            &            &            &            &              & \checkmark  \\
\hline
PSNR     &  27.94  & 28.34
  & 28.56
& 28.93 & 29.36
 & \textbf{29.46}    \\ 
SSIM     &  0.882  & 0.886
  & 0.887
 & 0.891 & 0.896
 & \textbf{0.897}     \\ \hline
\end{tabular}
 \vspace{0.4em}
\caption{Ablation result for SAPNet with different model (M) components. }
\label{Aba_Model}
\end{table}

\begin{table}[t]
\centering
\small
\begin{tabular}{l|lll}
\hline
Methods                                & Rain12      & Rain100L    & Rain100H                                       \\ \hline
Rainy                                  & 28.82/0.836 & 25.52/0.825 & 12.13/0.349                                 \\
DDN                                    & 28.89/0.897 & 26.25/0.856 & 12.65/0.420                              \\
RESCAN                                 & 33.60/0.953 & 31.76/0.946 & 27.43/0.841                                   \\ 
PreNet                                 & 34.79/\underline{0.964} & \textbf{36.09}/\underline{0.972} & 28.06/\underline{0.884}                            \\ 
Syn2Real       & 28.06/0.893 & 24.24/0.871 & 15.18/0.397                                 \\ 
MSPFN                                  & 34.17/0.945 & 30.55/0.915 & 26.29/0.798                               \\ 
MOSS           & 28.82/0.835 & 27.27/0.885 & 16.82/0.487                                 \\ 
EffDerain & 28.11/0.836 & 25.72/0.800 & 14.82/0.439          \\ 
MPRNet                                 & \textbf{36.53}/0.963 & 34.73/0.959 & \underline{28.52}/0.872                         \\ 
\textbf{Ours}           & \underline{35.50}/\textbf{0.968} & \underline{34.77}/\textbf{0.973} & \textbf{29.46/0.897}                       \\ \hline
\end{tabular}
\vspace{0.4em}
\caption{PSNR $\uparrow$ and SSIM $\uparrow$ comparison on Rain12, Rain100L and Rain100H}
\label{PSNR_SSIM}
\end{table}

\begin{table}[t]
\centering
\small
\begin{tabular}{l|lll}
\hline
Methods                                & Rain800     & SIRR                                & MOSS        \\ \hline
Rainy                                  & 0.755/26.63 & 0.672/29.13 & 0.786/26.47 \\ 
DDN                                    & 0.741/\textbf{18.12} & 0.670/25.46                         & 0.790/19.92 \\ 
RESCAN                                 & 0.761/21.54 & 0.671/25.67                         & 0.794/19.02 \\
PreNet                                 & \underline{0.762}/20.08 & 0.674/24.17                         & \underline{0.797}/18.26 \\
Syn2Real       & 0.750/\underline{20.04} & 0.689/24.11                         & 0.783/\underline{17.96} \\ 
MSPFN                                  & 0.749/22.17 & 0.657/\underline{20.71}                        & 0.732/22.64 \\ 
MOSS           & 0.743/22.05 & 0.691/29.06                         & 0.788/24.45 \\ 
EffDerain & 0.737/31.86 & 0.679/39.33                         & 0.773/38.10 \\ 
MPRNet                                 & 0.754/21.57 & \textbf{0.697}/28.48                         & \underline{0.797}/24.22 \\
\textbf{Ours}                                   & \textbf{0.767}/22.21 & \underline{0.696}/\textbf{20.68}                         & \textbf{0.798}/\textbf{17.88} \\ \hline
\end{tabular}
 \vspace{0.4em}
\caption{UNIQUE $\uparrow$ / BRISQUE $\downarrow$ comparison on Rain800, SIRR, and MOSS}
\label{UNIQUE_BRISQUE}
\end{table}

\begin{table}[t]
\centering
\tiny
\renewcommand\tabcolsep{2.5pt}
\scalebox{1.2}{
\begin{tabular}{l| l l l l l l l l l}
\hline
Metrics & Rainy & DDN  & RESCAN & PreNet  & EffDerain & Syn2Real & MOSS & \textbf{Ours} & GT   \\ \hline
mAP (\%)     & 52.1  & 65.1 & 78.5   & \underline{81.0}                            & 68.2           & 55.4                             & 73.2                         &   \textbf{82.2}             & 85.4 \\ 
mPA (\%)    & 65.3  & 66.4 & 70.3   & 73.8                            & 67.3           & 59.9                             & \underline{76.6}                        &  \textbf{77.2}             & 78.8 \\ 
mIOU (\%)   & 50.7  & 53.6 & 57.3   & 56.3                          & 56.7           & 49.9                             & \underline{60.1}                       & \textbf{62.2}              & 66.7 \\ \hline
\end{tabular}}
 \vspace{0.4em}
\caption{mAP$\uparrow$, mPA$\uparrow$ and mIOU$\uparrow$ comparison}
\label{mAP}
\end{table}

\begin{figure}[t]
\centering
\subfigure[Rainy]{
\includegraphics[width=2.5cm]{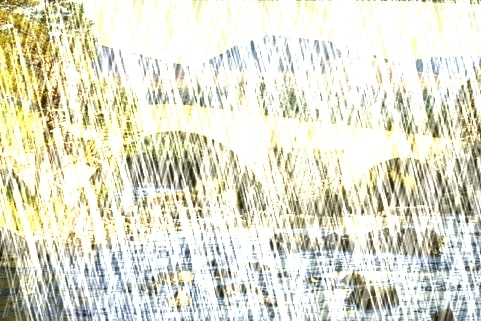}
} \hspace*{-0.9em}
\subfigure[RESCAN]{
\includegraphics[width=2.5cm]{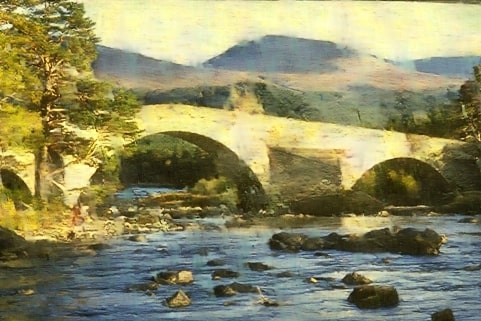}
} \hspace*{-0.9em}
\subfigure[PreNet]{
\includegraphics[width=2.5cm]{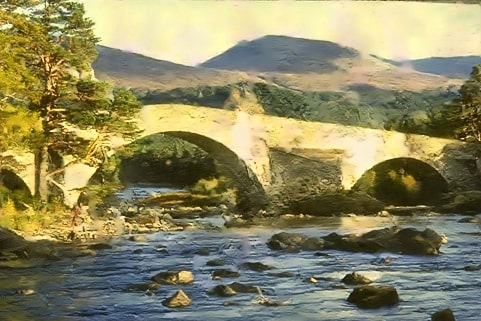}
} \hspace*{-0.9em} \\  \vspace{-0.9em}
\subfigure[MSPFN]{
\includegraphics[width=2.5cm]{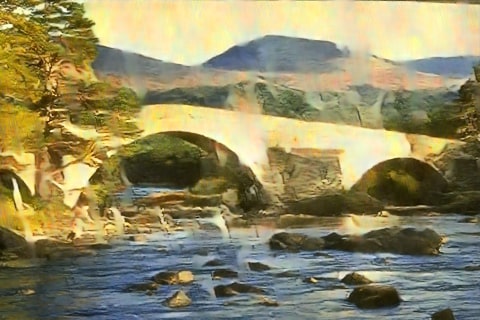}
} \hspace*{-0.9em}
\subfigure[Ours]{
\includegraphics[width=2.5cm]{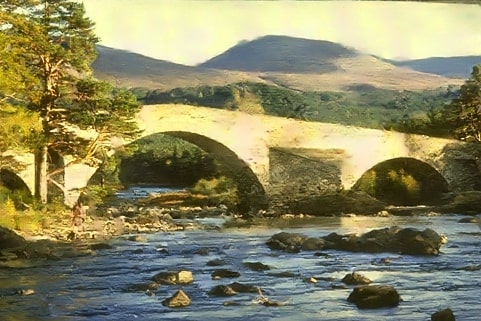}
} \hspace*{-0.9em}
\subfigure[GT]{
\includegraphics[width=2.5cm]{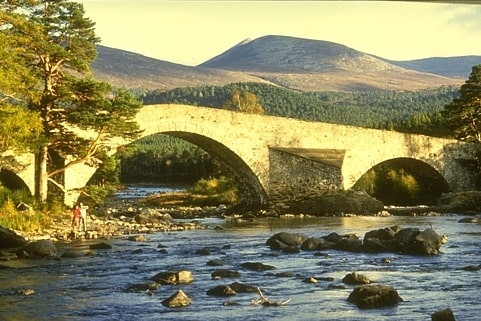}
} \hspace*{-0.9em}
\caption{Visual comparison at Rain100H}
\label{Rain100H(2)}
\end{figure}

\begin{figure}[t]
\centering
\subfigure[Rainy]{
\includegraphics[width=2.5cm]{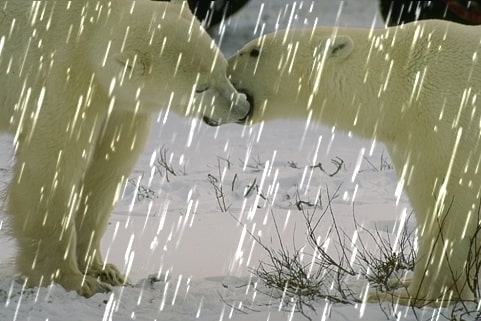}
} \hspace*{-0.9em}
\subfigure[DDN]{
\includegraphics[width=2.5cm]{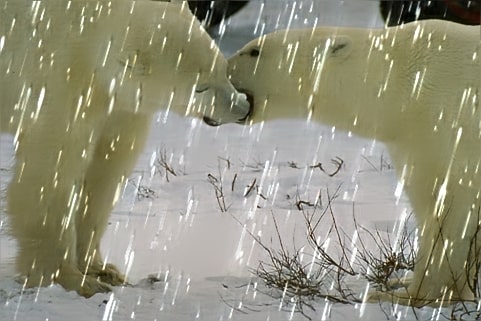}
} \hspace*{-0.9em}
\subfigure[MSPFN]{
\includegraphics[width=2.5cm]{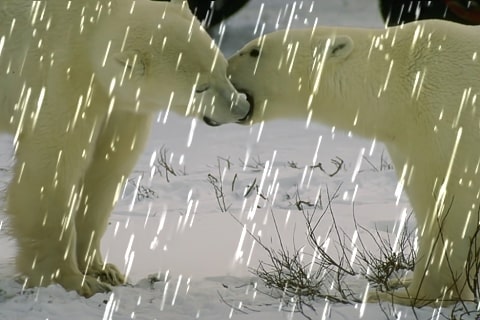}
} \hspace*{-0.9em} \\  \vspace{-0.9em}
\subfigure[MPRNet]{
\includegraphics[width=2.5cm]{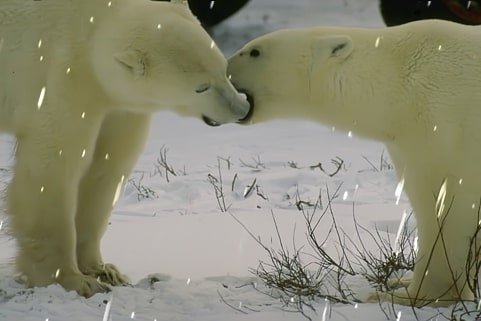}
} \hspace*{-0.9em}
\subfigure[Ours]{
\includegraphics[width=2.5cm]{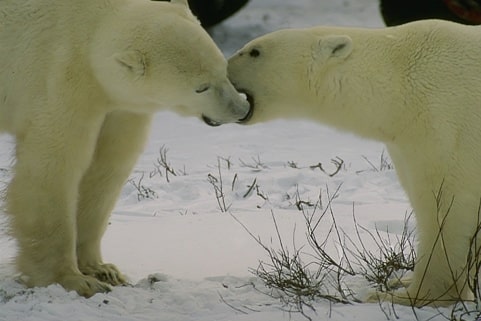}
} \hspace*{-0.9em}
\subfigure[GT]{
\includegraphics[width=2.5cm]{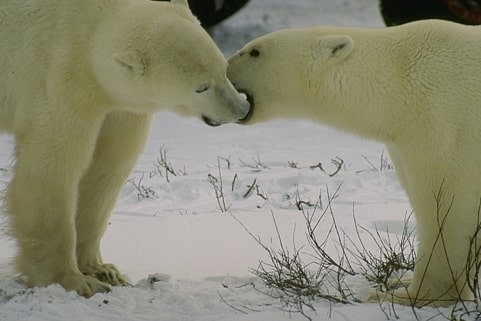}
} \hspace*{-0.9em}
\caption{Visual comparison at Rain100L}
\label{Rain100L}
\end{figure}		
		
\subsection{Ablation Study}
We conduct comprehensive ablation studies to investigate the contributions of each component to SAPNet's rain removal performance. The ablation studies are evaluated on synthetic rainy datasets due to the requirements of PSNR and SSIM.

\noindent
\textbf{Ablation of Attention Blocks}
The first ablation study aims to demonstrate the superiority of the proposed channel residual attention (CRA) block compared with the squeeze-excitation (SE) block and channel attention (CA) block. Table \ref{Aba_attention} shows the PSNR and SSIM for SAPNet with different attention blocks. We note that SAPNet-CRA has the best PSNR and SSIM with an efficient inference time. Fig. \ref{VisAttention} displays the corresponding visual comparison. We can see that both SAPNet-Conv and SAPNet-SE fail to clear rain streaks in the sky. Although SAPNet-CA effectively removes large rain streaks, it over-smooth the backgrounds. In comparison, SAPNet-CRA preserves the most background textures and has the most pleasant looking.

\noindent
\textbf{Ablation of Model Components}
The second ablation study examines the effectiveness of different model components for SAPNet. Table \ref{Aba_Model} shows different versions of SAPNet, where we sequentially add channel residual attention (CRA), unsupervised background segmentation (UBS), perceptual contrastive loss (PCL), dilation, learning rate decay and learned perceptual image similarity loss (LPISL). We notice that each component contributes to better rain removal (i.e., better PSNR and SSIM). The visual comparison in Fig. \ref{VisModel} also demonstrates that the proposed modules help rain removal and facial details preservation.

\noindent
\textbf{Ablation of Contrastive Losses}
The third ablation study investigates the effectiveness of the proposed perceptual contrastive loss (PCL). Table \ref{CompareContrastLimit} compares SAPNet's performance with no contrastive loss, L1 contrastive loss, and perceptual contrastive loss. We can see that L1 contrastive loss significantly degrade the rain removal performances, whereas the perceptual contrastive loss substantially improves the performances. We also make a visual comparison in Fig. \ref{VisContrast}, with PreNet as an additional reference. It shows that SAPNet with no contrastive loss or with L1 contrastive loss fails in large and long rain streaks. In comparison, SAPNet with the proposed PCL successfully remove different types of rain streaks.



\subsection{Comparison on Synthetic Rainy Dataset}
We make a quantitative comparison for synthetic rainy datasets in Table \ref{PSNR_SSIM}. It can be seen that SAPNet has the best SSIM for all, and the second-best PSNR for Rain12 and Rain100L. For the most challenging Rain100H, SAPNet has the best PSNR and SSIM. We also conduct a visual comparison on synthetic rainy images. Fig \ref{Rain100H(2)} (Rain100H) shows that RESCAN, PreNet, and MSPFN clear most heavy rain streaks but leave significant grey marks on the background sky. In comparison, SAPNet has the best rain removal performance and is closest to the groundtruth. Fig \ref{Rain100L} (Rain100L) shows that DDN and MSPFN fail to clear the long rain streaks and that MPRNet's local details are unpromising. In contrast, SAPNet's derained image is almost on par with the groundtruth.

\subsection{Comparison on Real-World Rainy Images}
\begin{figure}[t]
\centering
\subfigure[Rainy]{
\includegraphics[width=2.0cm]{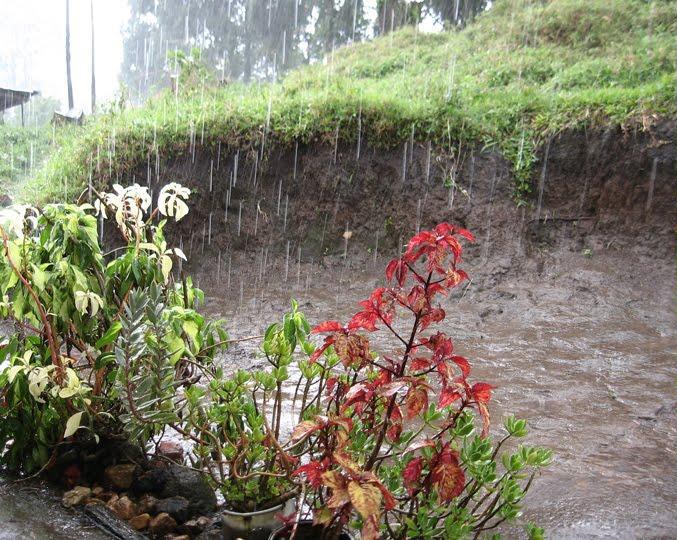} 
} \hspace*{-0.9em}
\subfigure[DDN]{
\includegraphics[width=2.0cm]{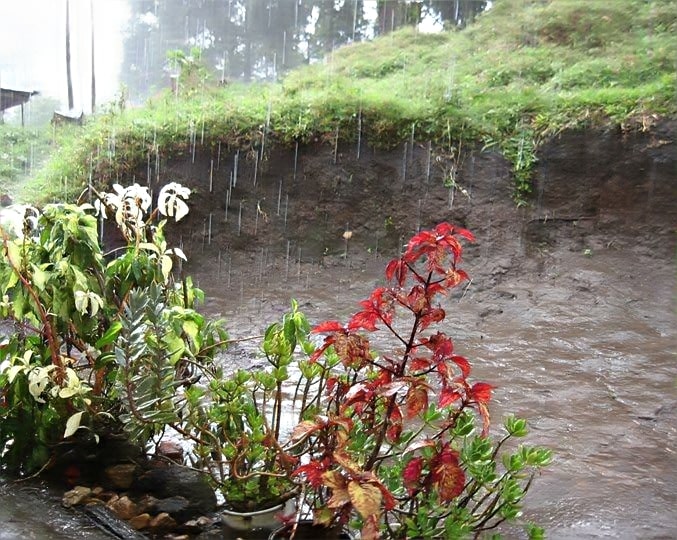}
} \hspace*{-0.9em}
\subfigure[Syn2Real]{
\includegraphics[width=2.0cm]{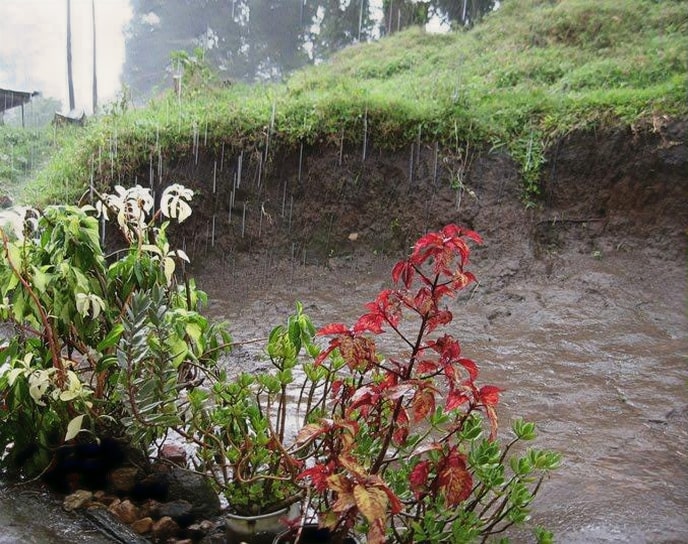}
} \hspace*{-0.9em}
\subfigure[MSPFN]{
\includegraphics[width=2.0cm]{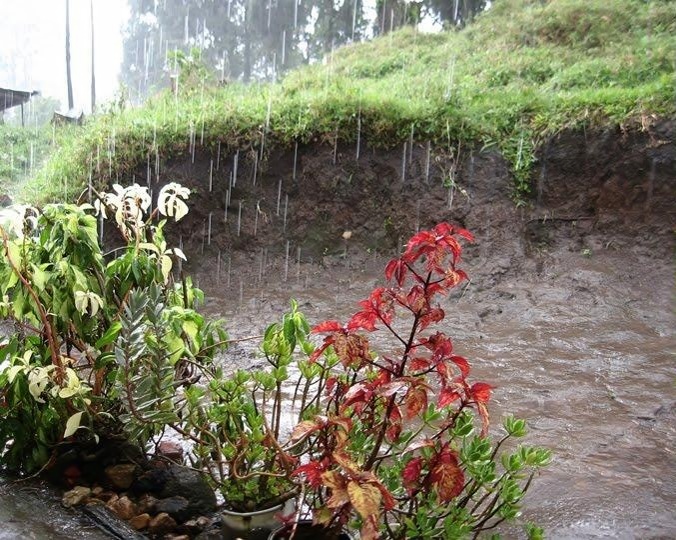}
} \hspace*{-0.9em} \\  \vspace{-0.9em}
\subfigure[MOSS]{
\includegraphics[width=2.0cm]{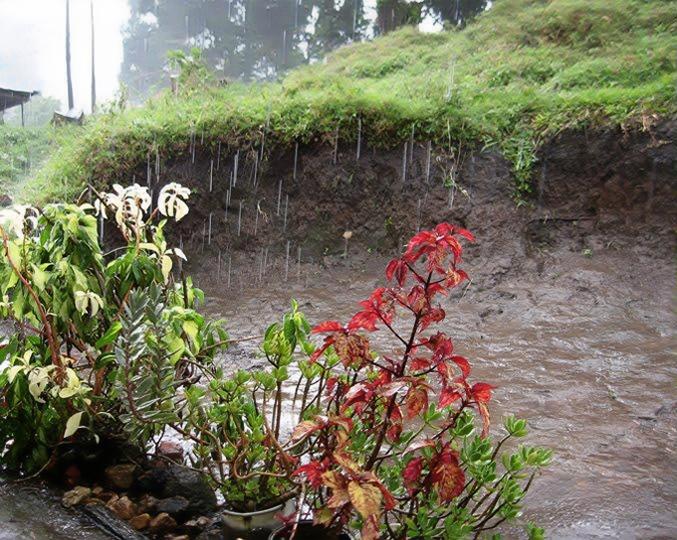}
} \hspace*{-0.9em}
\subfigure[EffDerain]{
\includegraphics[width=2.0cm]{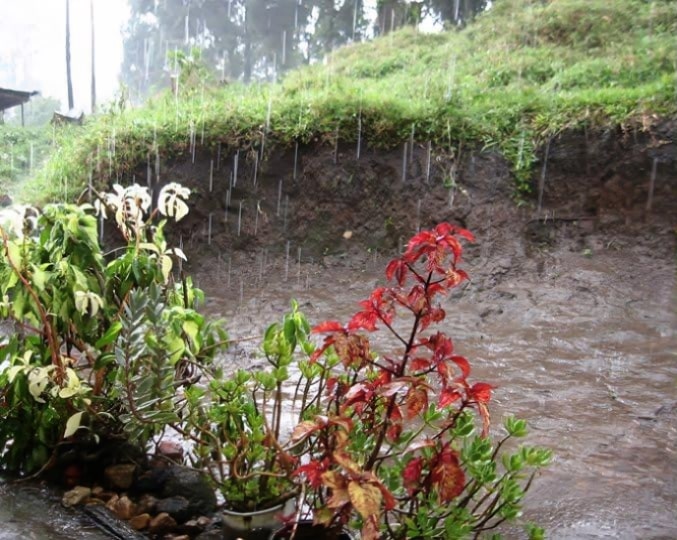}
} \hspace*{-0.9em}
\subfigure[MPRNet]{
\includegraphics[width=2.0cm]{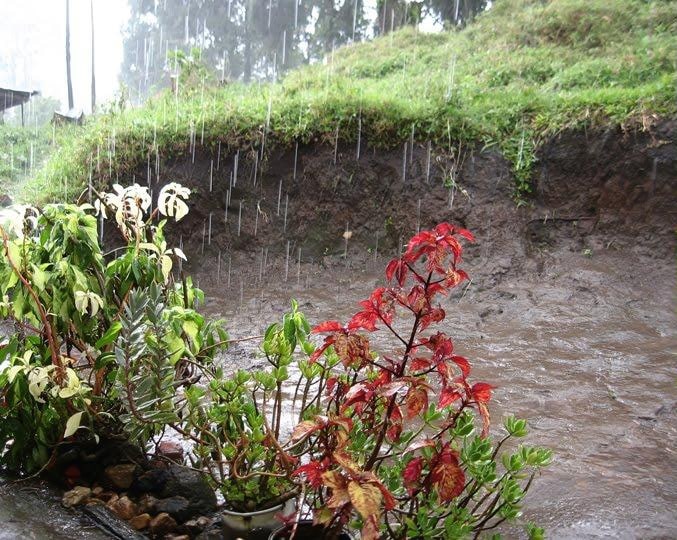}
} \hspace*{-0.9em}
\subfigure[Ours]{
\includegraphics[width=2.0cm]{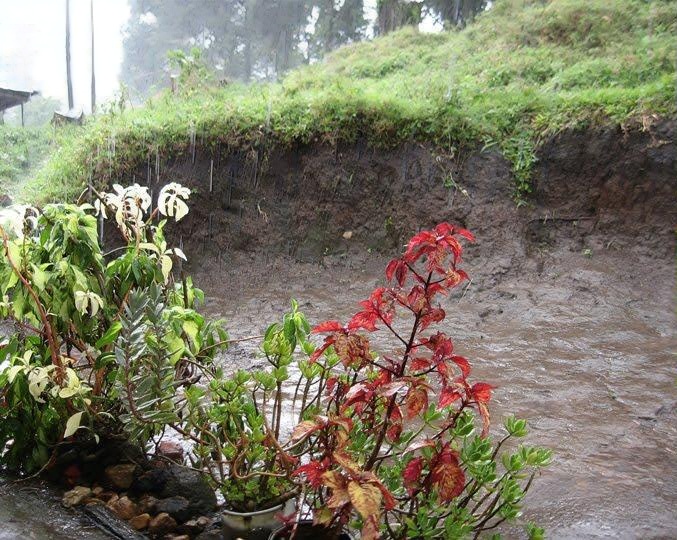}
} \hspace*{-0.9em} \\  
\caption{Visual comparison at Rain800}
\label{Real_Rain800}
\end{figure}

\begin{figure}[t]
\centering
\subfigure[Rainy]{
\includegraphics[width=2cm]{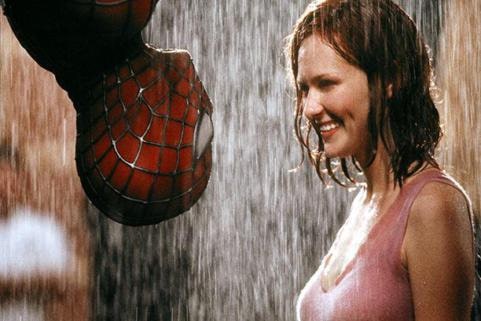}
} \hspace*{-0.9em}
\subfigure[DDN]{
\includegraphics[width=2cm]{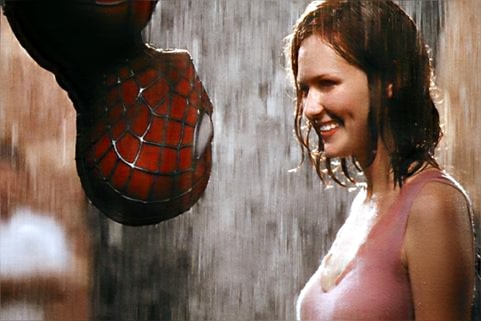}
} \hspace*{-0.9em}
\subfigure[RESCAN]{
\includegraphics[width=2cm]{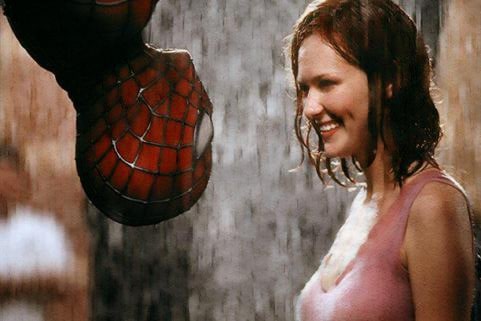}
} \hspace*{-0.9em}
\subfigure[MSPFN]{
\includegraphics[width=2cm]{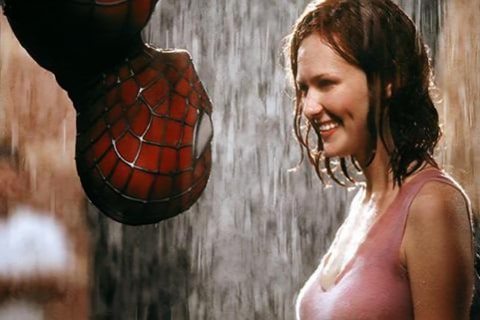}
} \hspace*{-0.9em} \\  \vspace{-0.9em}
\subfigure[EffDerain]{
\includegraphics[width=2cm]{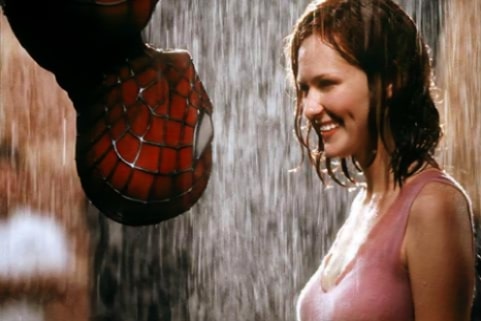}
} \hspace*{-0.9em}
\subfigure[MOSS]{
\includegraphics[width=2cm]{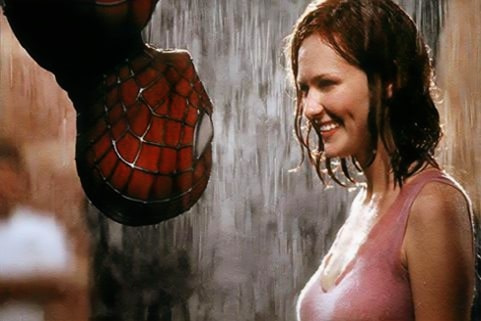}
} \hspace*{-0.9em}
\subfigure[MPRNet]{
\includegraphics[width=2cm]{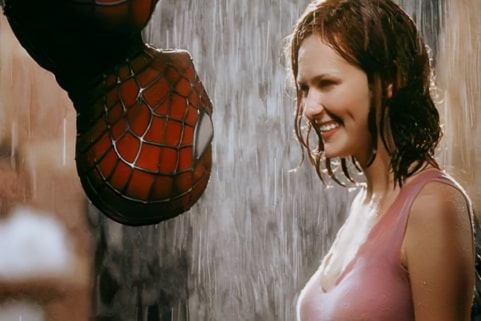}
} \hspace*{-0.9em}
\subfigure[Ours]{
\includegraphics[width=2cm]{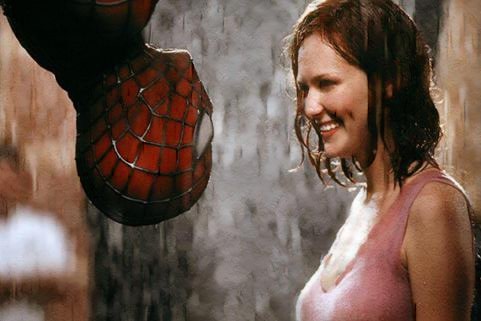}
} \hspace*{-0.9em} \\  
\caption{Visual comparison at SIRR}
\label{Real_SIRR_1}
\end{figure}

We make a quantitative comparison for real-world rainy datasets in Table \ref{UNIQUE_BRISQUE}. It shows that SAPNet has the best BRISQUE for SIRR and the best UNIQUE for Rain800. For the recently proposed MOSS dataset, SAPNet outperforms all competing models in terms of UNIQUE and BRISQUE. We also conduct qualitative comparisons on Rain800 (Fig. \ref{Real_Rain800}) and SIRR (Fig. \ref{Real_SIRR_1}). It shows that other methods (1) fail to clear the rain streaks (2) introduce blur and under/over-exposure. In contrast, SAPNet maintains the best brightness and exposure while removing diverse types of rain streaks effectively.


\begin{figure}[t]
\centering
\subfigure[Rainy]{
\includegraphics[width=2cm]{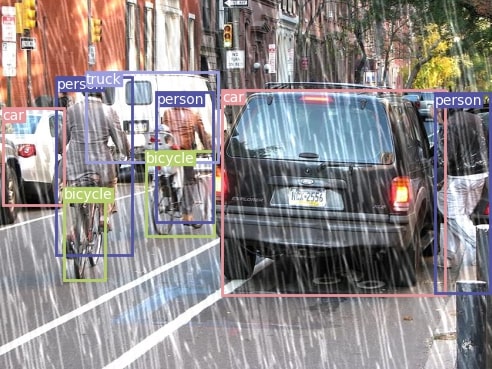}
} \hspace*{-0.9em}
\subfigure[DDN]{
\includegraphics[width=2cm]{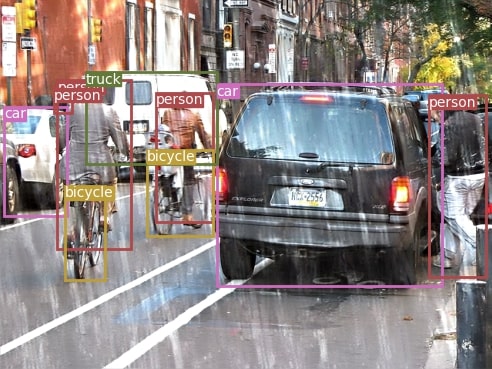}
} \hspace*{-0.9em}
\subfigure[Syn2Real]{
\includegraphics[width=2cm]{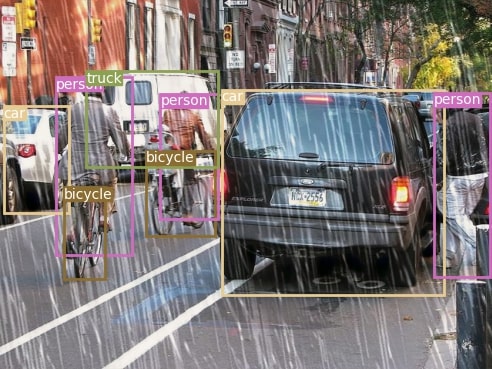}
} \hspace*{-0.9em}
\subfigure[MOSS]{
\includegraphics[width=2cm]{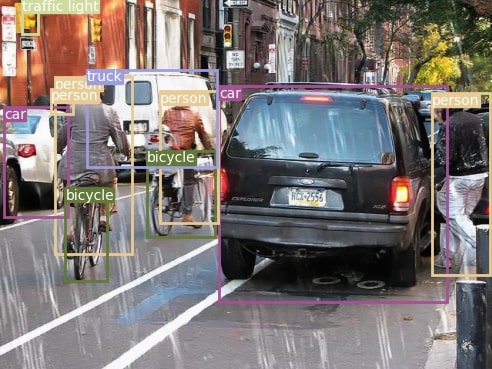}
} \hspace*{-0.9em} \\ \vspace{-0.9em}
\subfigure[EffDerain]{
\includegraphics[width=2cm]{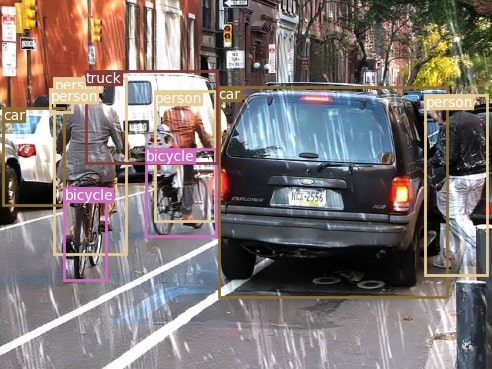}
} \hspace*{-0.9em}
\subfigure[RESCAN]{
\includegraphics[width=2cm]{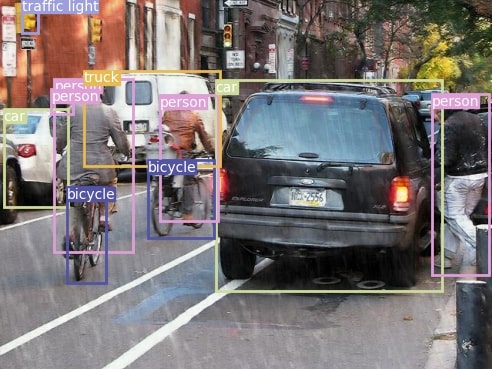}
} \hspace*{-0.9em}
\subfigure[Ours]{
\includegraphics[width=2cm]{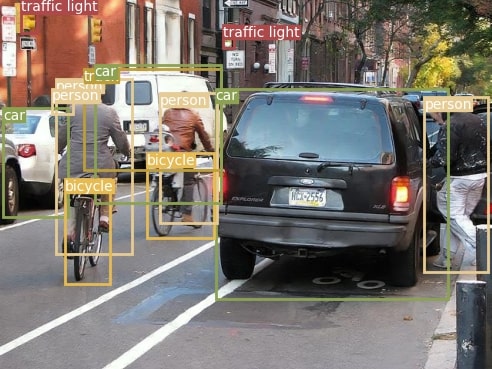}
} \hspace*{-0.9em}
\subfigure[GT]{
\includegraphics[width=2cm]{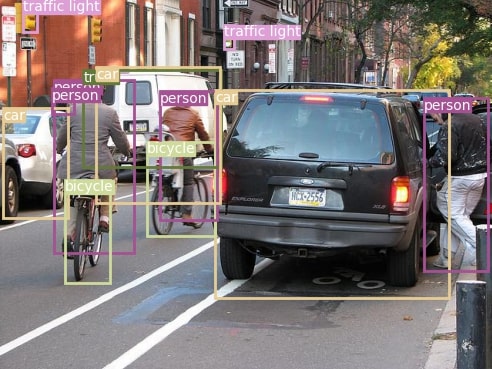}
} \hspace*{-0.9em} \\  
\caption{Object detection result at COCO150}
\label{Detection}
\end{figure}

\begin{figure}[t]
\centering
\subfigure[Rainy]{
\includegraphics[width=1.6cm]{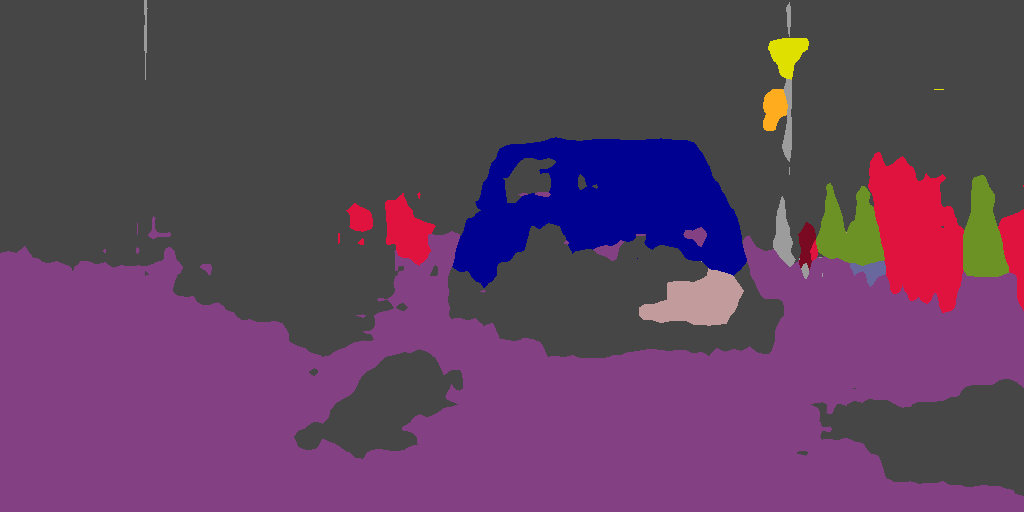}
}\hspace*{-0.5em}
\subfigure[DDN]{
\includegraphics[width=1.6cm]{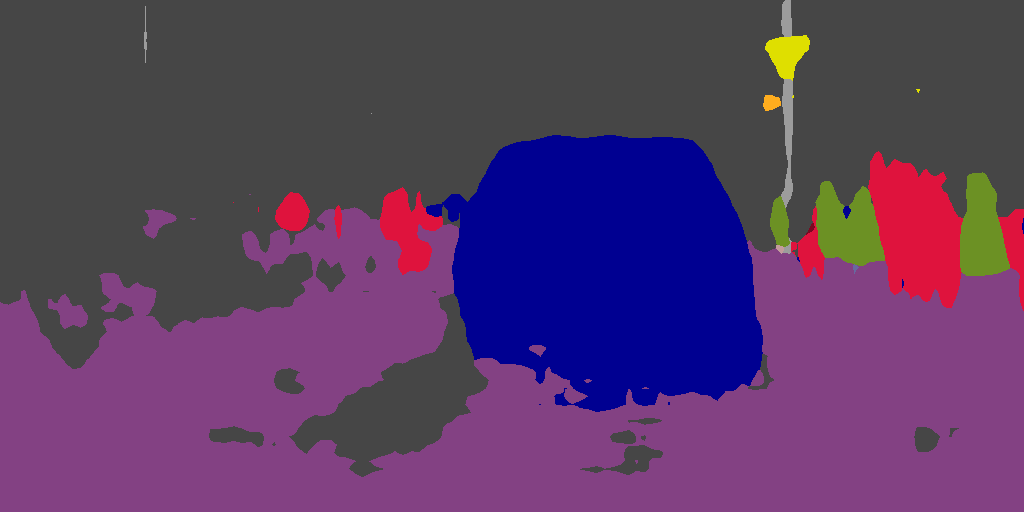}
}\hspace*{-0.5em}
\subfigure[Syn2Real]{
\includegraphics[width=1.6cm]{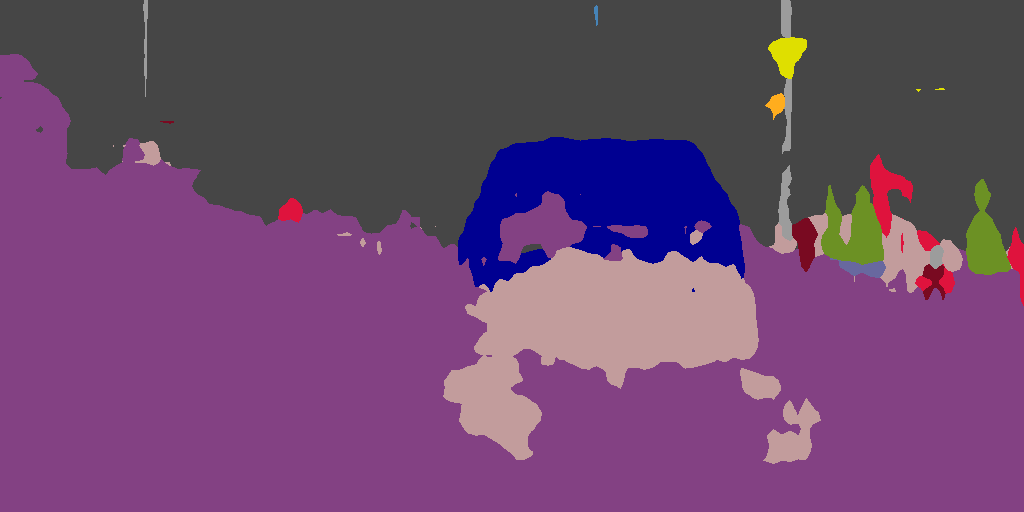}
}\hspace*{-0.5em}
\subfigure[MOSS]{
\includegraphics[width=1.6cm]{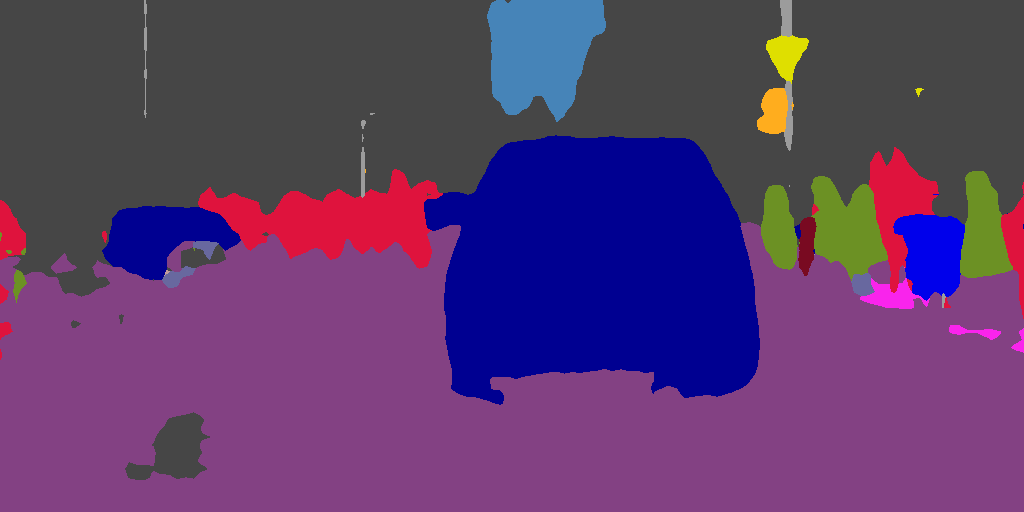}
}\hspace*{-0.5em}
\subfigure[EffDerain]{
\includegraphics[width=1.6cm]{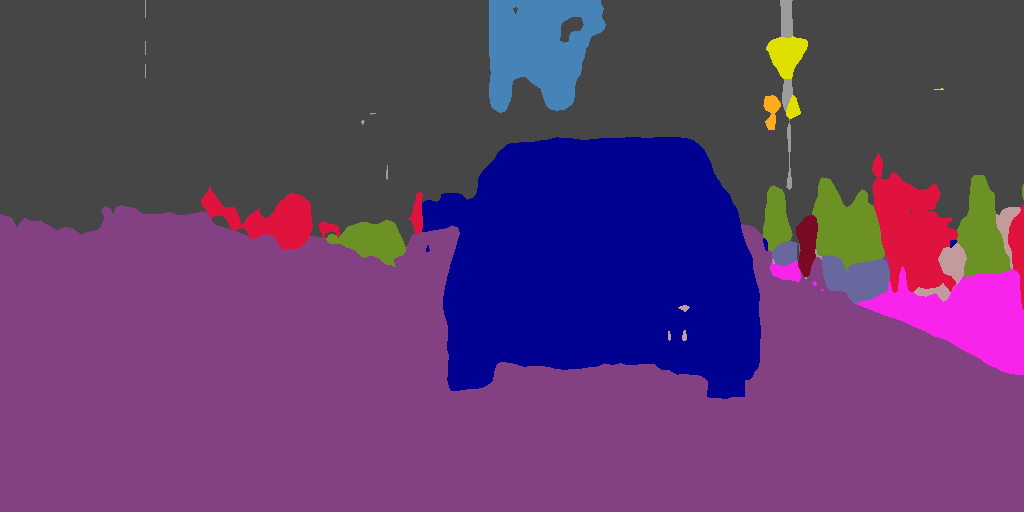}
}                    \\ \vspace{-0.9em}
\subfigure[RESCAN]{
\includegraphics[width=1.6cm]{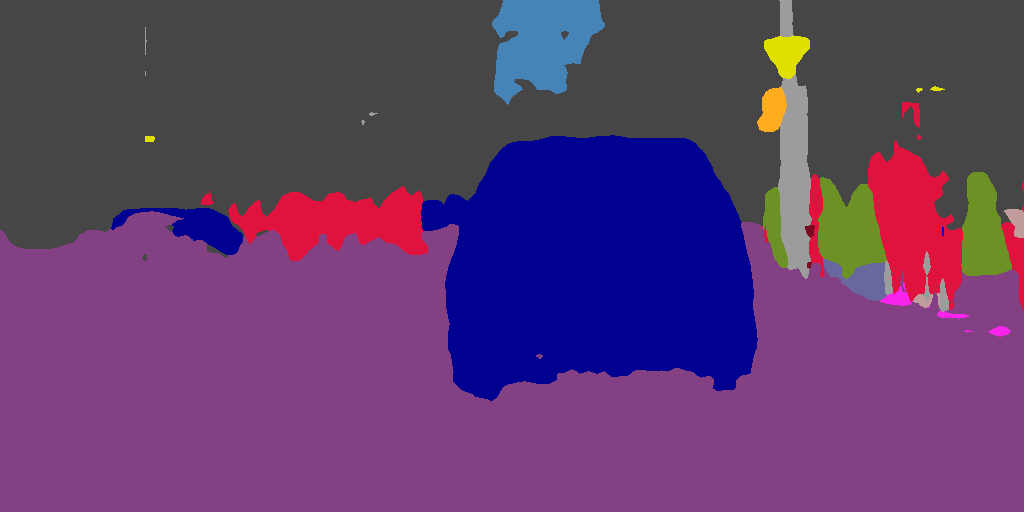}
}\hspace*{-0.5em}
\subfigure[PreNet]{
\includegraphics[width=1.6cm]{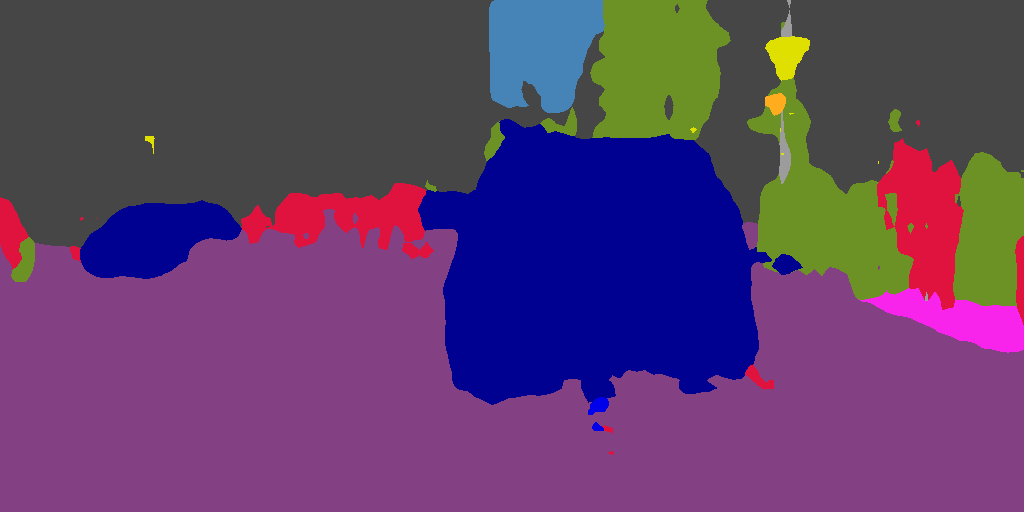}
}\hspace*{-0.5em}
\subfigure[MSPFN]{
\includegraphics[width=1.6cm]{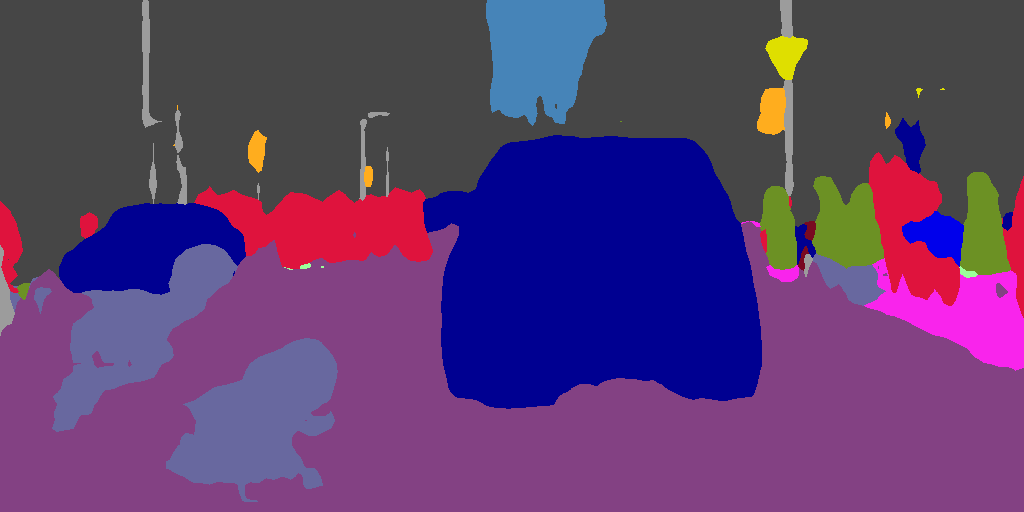}
}\hspace*{-0.5em}
\subfigure[Ours]{
\includegraphics[width=1.6cm]{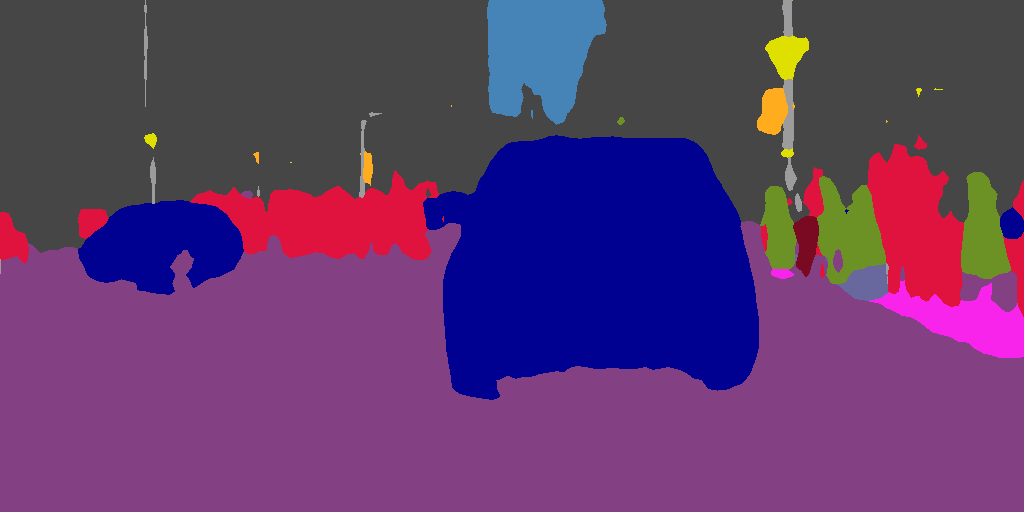}
}\hspace*{-0.5em}
\subfigure[GT]{
\includegraphics[width=1.6cm]{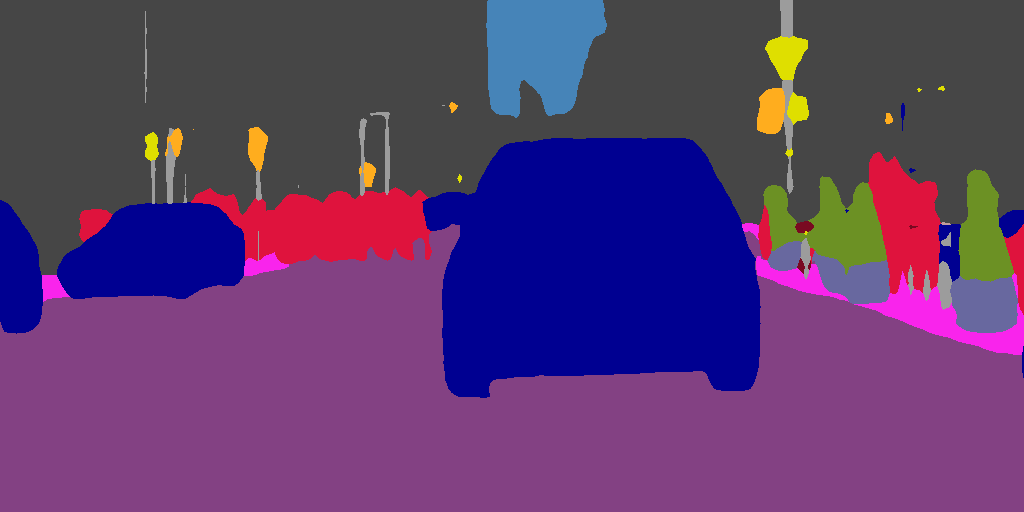}
} \hspace*{-0.5em} \\  
\caption{Semantic segmentation result at CityScape150}
\label{Segmentation}
\end{figure}

\subsection{Model Efficiency}
Real-time deraining on mobile devices demands an affordable model size with a fast inference speed. It is essential to investigate the model efficiency regarding the number of parameters and the inference time (Fig. \ref{TimeEfficiency}). More details will be in the supplementary material.

\subsection{Detection and Segmentation}
To investigate the contribution of deraining models to high-level vision tasks, we use Yolov3 \cite{redmon2018yolov3} for object detection on COCO150 and PSPNet \cite{zhao2017pyramid} for semantic segmentation on CityScape 150. 

Fig. \ref{Detection} reveals that the competing models have limited ability to conduct rain removal for object detection. For example, DDN, Syn2Real, and EffDerain fail to detect the traffic lights, which could be disastrous for automatic pilots. In contrast, SAPNet removes most rain streaks and helps bridge the detection to the groundtruth. 

Fig. \ref{Segmentation} displays that the deraining method for comparison also has limited contribution to semantic segmentation. For instance, DDN, Syn2Real, EffDerain, and RESCAN miss the left car in the segmentation map. In comparison, SAPNet has the most accurate segmentation and is closest to the groundtruth. The quantitative comparisons in Table \ref{mAP} further demonstrates that SAPNet has the best performance in both object detection and semantic segmentation.

\section{Conclusion}
This paper presented a segmentation-aware progressive network for image deraining. Firstly, we designed a progressive dilated unit (PDU) to utilize the multi-scale rain streaks information. Secondly, we proposed perceptual contrastive loss (PCL) and learned perceptual image similarity loss (LPISL) to bridge the derained image to the groundtruth in terms of pixel-wise and perceptual-level differences. Finally, we leveraged unsupervised background segmentation (UBS) to reserve the semantic information during exhaustive rain removal. Extensive experiments demonstrate the effectiveness of the proposed method. Our future work will explore detection-driven deraining and investigate rain removal at sub-optimal illumination.


\balance

{\small
\bibliographystyle{ieee_fullname}
\bibliography{wacv}
}

\end{document}